\documentclass{article}
\usepackage{iclr2027_conference,times}
\usepackage[T1]{fontenc}
\usepackage[utf8]{inputenc}
\usepackage{graphicx}
\usepackage{amsmath,amssymb,booktabs,array,tabularx,multirow,graphicx}
\usepackage[table]{xcolor}
\definecolor{cwLav}{HTML}{DDD6E5}\definecolor{cwLavL}{HTML}{DFEDF2}\definecolor{cwGreyL}{HTML}{BFDCE7}
\definecolor{cwPeachL}{HTML}{F5D8BF}\definecolor{cwPeach}{HTML}{D9E4C2}\definecolor{cwCoral}{HTML}{CAC2D7}\definecolor{cwBg}{HTML}{F9EDE0}
\usepackage{hyperref,url}
\hypersetup{colorlinks=true,citecolor=blue,linkcolor=blue,urlcolor=blue,pdfauthor={},pdftitle={Joint Driver, Vehicle, and Road Modeling for Forecasting and Driver Monitoring}}
\newcommand{\model}{TriDrive}
\newcommand{\sd}[2]{\ensuremath{#1\!\pm\!#2}}
\newcommand{\sig}{\ensuremath{^{\ast}}}

\newcolumntype{L}[1]{>{\raggedright\arraybackslash}p{#1}}
\title{Joint Driver, Vehicle, and Road 
Modeling for Forecasting and Driver Monitoring}
\iclrfinalcopy   
\author{Yuhang Wang$^{1}$, Jingxin Yang$^{2}$, Chuheng Wei$^{3}$, Yuechen Guo$^{1}$, \\
\bfseries Jinghan Xu$^{4}$, Zhao Han$^{1}$, Hao Zhou$^{1}$ \\[3pt]
{\mdseries $^{1}$University of South Florida \quad $^{2}$NVIDIA \quad $^{3}$Purdue University \quad $^{4}$Hunan University}}
\newcommand{\gsd}[2]{#1\,{\scriptsize\textcolor{black!55}{$\pm$#2}}}
\begin{document}
\maketitle
\begin{center}\vspace{-14pt}Code, checkpoints, and demo video: \url{https://huggingface.co/HenryYHW/TriDrive}\vspace{2pt}\end{center}
\suppressfloats[t]
\begin{abstract}
Predicting how drivers, vehicles, and road scenes interact and evolve together is central to driver monitoring.
Prior work models in-cabin activity or traffic-conditioned driver motion in isolation, motivating joint driver, vehicle, and road modeling with real-time on-vehicle evaluation.
We introduce \model{}, to our knowledge the first unified framework that jointly forecasts driver kinematics, vehicle dynamics, and road demands through an automation-conditioned transition model.
Modality-specific encoders (an anchored kinematic representation of the driver, causal CAN-bus dynamics, and frozen V-JEPA~2 road latents with structured road margins) are connected by directed residual connections through which driver and road context refine vehicle forecasts.
We evaluate \model{} on three downstream tasks.
On the public AIDE benchmark, its kinematic encoder recipe sets a new full-set state of the art (SOTA) among published baselines (48.05 versus 71.47 All-MPJPE).
On 197.2 hours of naturalistic BATON subset, directed connections and road margins raise assistance-engaged PR-AUC by 0.084 for steering onset and 0.286 for time-to-collision drops.
For real-time use, we distill the road encoders and run \model{} on a comma four with an external 8\,GB GPU, where a lightweight current-state warning probe updates at 5\,Hz with 177\,ms p95 latency while the joint model forecasts concurrently.
The probe is above an openpilot-based baseline on human-labeled manual-driving warnings (AUROC 0.725 versus 0.563), and in a paired on-road study 14 drivers rate its warnings as more appropriate (+1.79) and timely (+2.67) than those of openpilot's driver-monitoring system.
\end{abstract}

\section{Introduction}
\label{sec:intro}

\begin{figure}[htbp]
    \centering
    \includegraphics[width=\linewidth]{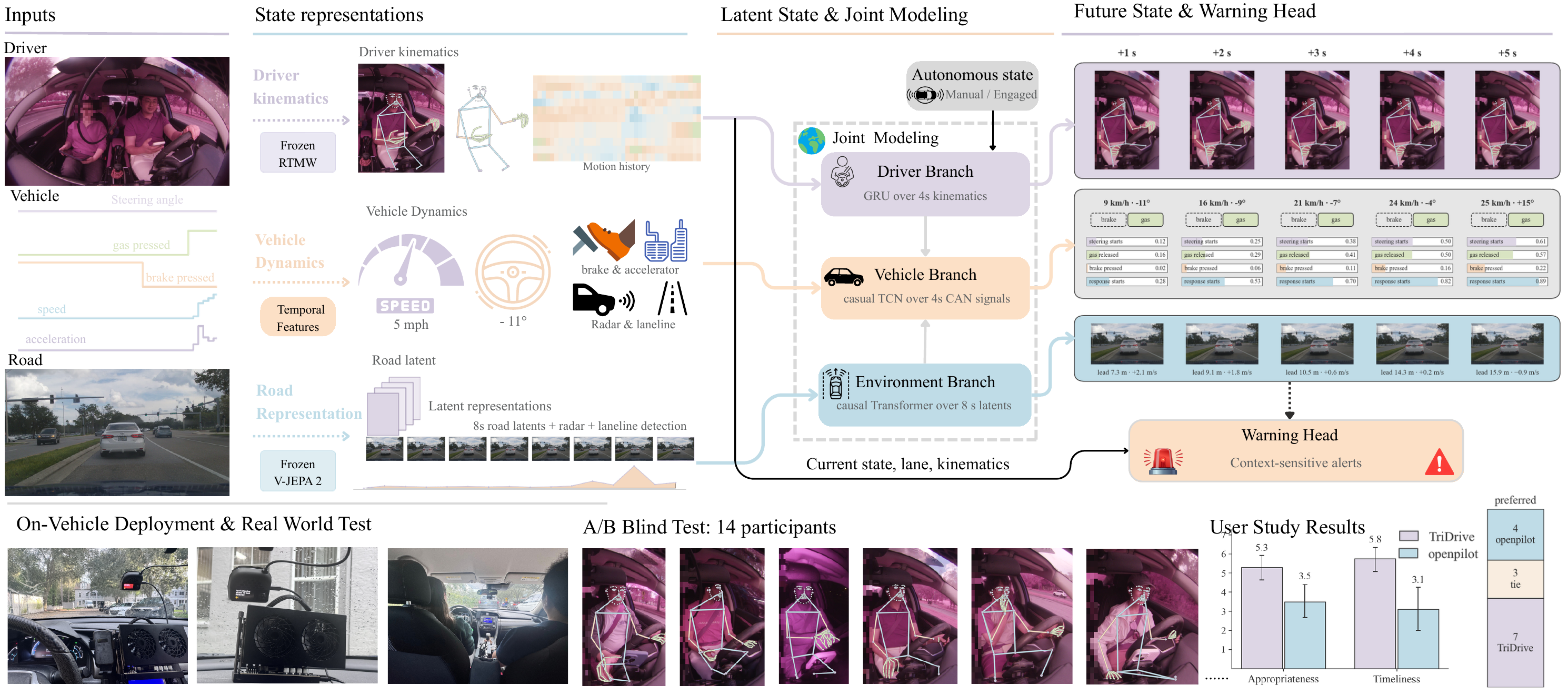}
    \caption{
    \textbf{Overview of \model{}.}
    \textbf{Top:} Driver kinematics, vehicle dynamics, and road latents from a frozen V-JEPA~2 \citep{assran2025} encoder support automation-conditioned joint forecasting (slow horizon).
    \textbf{Bottom:} On-vehicle deployment and a paired user study comparing \model{} with openpilot  \citep{commaai2026openpilot} and chestnut  \citep{commaai2026chestnut}, involving 14 participants (8.4 hours).
    }
    \label{fig:teaser}
\end{figure}

A brief glance at a phone while stationary at a red light may not warrant an immediate warning. If the driver continues looking down as the light turns green or the vehicle creeps toward the car ahead, however, a warning may become appropriate. Driver-state recognition alone cannot distinguish these situations. Road context must also be interpreted together with vehicle dynamics to determine what requires the driver's attention and how soon. These dependencies motivate joint understanding and prediction of driver, vehicle, and road states during both manual and assisted driving.

Existing work addresses different parts of this problem. Driver activity recognition estimates current behavior from cabin observations \citep{martin2019,ortega2020,kopuklu2021}. Maneuver anticipation combines interior and exterior cues to predict upcoming actions \citep{jain2016,ma2023}. Driver-WM forecasts in-cabin dynamics conditioned on traffic context \citep{chi2026}, while driver-state world modeling supports selective multimodal monitoring \citep{qiu2026}. None of these systems forecasts driver motion, vehicle dynamics, and road demands together, and none has been run in real time on a vehicle and evaluated with drivers.

We introduce \model{}, a unified framework that jointly models driver, vehicle, and road states for forecasting and driver monitoring.
In Figure~\ref{fig:teaser}, a frozen pose estimator extracts driver kinematics, CAN-bus histories describe vehicle dynamics, and a frozen V-JEPA~2 encoder produces road latents augmented with road-state signals. Three separately pretrained temporal predictors learn the evolution of these structured and latent states conditioned on automation state. Directed residual coupling lets driver and road refine vehicle forecasts.

The framework runs on two horizons.
The joint model forecasts state and hazard transitions over the next five seconds (slow horizon), while a separate lightweight probe reads the current driver, vehicle, and road state to issue warnings at 5\,Hz (fast horizon).
Separating the probe keeps its inputs interpretable. Section~\ref{sec:warning} shows that adding forecast features does not consistently improve offline warning AUROC, so the deployed probe reads current states.

Real-time deployment introduces additional computational constraints. A large road encoder can delay cabin processing when both share a vehicle-mounted GPU. We distill the road encoder and road-risk model into a compact student while preserving their downstream interfaces. Cooperative GPU scheduling, reduced module rates, and a staleness fallback let it run next to openpilot. We verify the deployed configuration with replay comparisons and on-vehicle timing.

We evaluate \model{} on three tasks: in-cabin kinematic forecasting on AIDE \citep{yang2023}, where its kinematic recipe reaches the lowest published All-MPJPE; cross-domain dynamics anticipation on 197.2 hours of BATON \citep{wang2026}; and a real-time contextual warning probe, assessed on human-labeled warnings, on the vehicle, and in a paired on-road study.
In that study, 14 participants drive their own vehicles with both systems over 8.4 hours of dual-device recording and rate \model{}'s warnings as more appropriate and timely than those of openpilot's driver-monitoring system, and as more annoying.

Our contributions are fourfold:\\
1.~\textbf{Joint driver, vehicle, and road modeling.} \model{} connects three pretrained predictors through directed residuals. On BATON, these connections and structured road margins improve assistance-engaged PR-AUC by 0.084 for steering onset and 0.286 for TTC drops, respectively. Both gains reproduce on 60 additional drivers.\\
2.~\textbf{Modality-specific representations.} We use anchored driver kinematics with BATON pose pretraining and fuse V-JEPA~2 road latents with structured road margins. The kinematic recipe sets a new full-set state of the art on AIDE (48.05 versus 71.47 All-MPJPE).\\
3.~\textbf{Edge deployment.} Distilled road encoders and cooperative GPU scheduling enable joint forecasting to run alongside a 5\,Hz current-state warning probe (177\,ms p95) on comma four and chestnut.\\
4.~\textbf{Downstream validation with drivers.} The warning probe is above an openpilot-based baseline on human-labeled manual-driving warnings, and 14 participants rate it as more appropriate and timely than openpilot's driver-monitoring system in a paired on-road study.

\section{Related work}
\label{sec:related}

\textbf{Datasets and benchmarks.}
Drive\&Act, DMD, and DAD support driver activity recognition and anomaly detection \citep{martin2019,ortega2020,kopuklu2021}. AIDE provides a comprehensive training and evaluation resource with synchronized cabin and road views and rich annotations \citep{yang2023}. However, limited large-scale synchronized data constrain joint driver--vehicle--road modeling. BATON's growing naturalistic corpus combines cabin and road videos with vehicle signals, enabling broader training and evaluation \citep{wang2026}. DriveMotion further standardizes driver motion forecasting across multiple sources \citep{wang2026drivemotion}. We primarily train and evaluate on AIDE and BATON.

\textbf{Driver monitoring and forecasting.}
Brain4Cars and CEMFormer combine in-cabin observations with external context for maneuver anticipation \citep{jain2016,ma2023}. Driver-WM forecasts driver motion conditioned on traffic context \citep{chi2026}. Risk-aware driver-state modeling combines visual and physiological signals with predictive features for selective monitoring \citep{qiu2026}. PV-WM jointly predicts pedestrian and vehicle states from structured tracks \citep{chi2026pvwm}. These methods model in-cabin drivers or external traffic agents without jointly forecasting driver, vehicle, and road states. Meanwhile, openpilot's deployed driver monitor focuses on driver attention assessment \citep{commaai2026openpilot}. \model{} jointly forecasts all three domains and treats driver monitoring as one downstream task, evaluated with a real-time on-vehicle warning probe rated by participants.

\textbf{Motion representations and distillation.}
Graph networks \citep{yan2018} and MLPs \citep{guo2023} model skeletal sequences, while MotionBERT uses a spatiotemporal Transformer pretrained to recover 3D motion from noisy, partial 2D observations \citep{zhu2023}. Video VAEs learn compact representations through reconstruction \citep{videovae}, while V-JEPA~2 learns predictive video latents \citep{assran2025}. We combine driver kinematics with frozen road representations for joint forecasting and evaluate alternative motion encoders. V-JEPA~2.1 distills predictive representations from a frozen teacher into smaller video encoders \citep{murlabadia2026vjepa21}. Following the teacher--student principle \citep{hinton2015}, we distill road features and risk outputs into a compact student for on-vehicle execution.

\section{\model{}: representations and joint modeling}
\label{sec:method}


\subsection{Problem formulation: co-evolution of driver, vehicle, and road}

\begin{figure}[htbp]
    \centering
    \includegraphics[width=\linewidth]{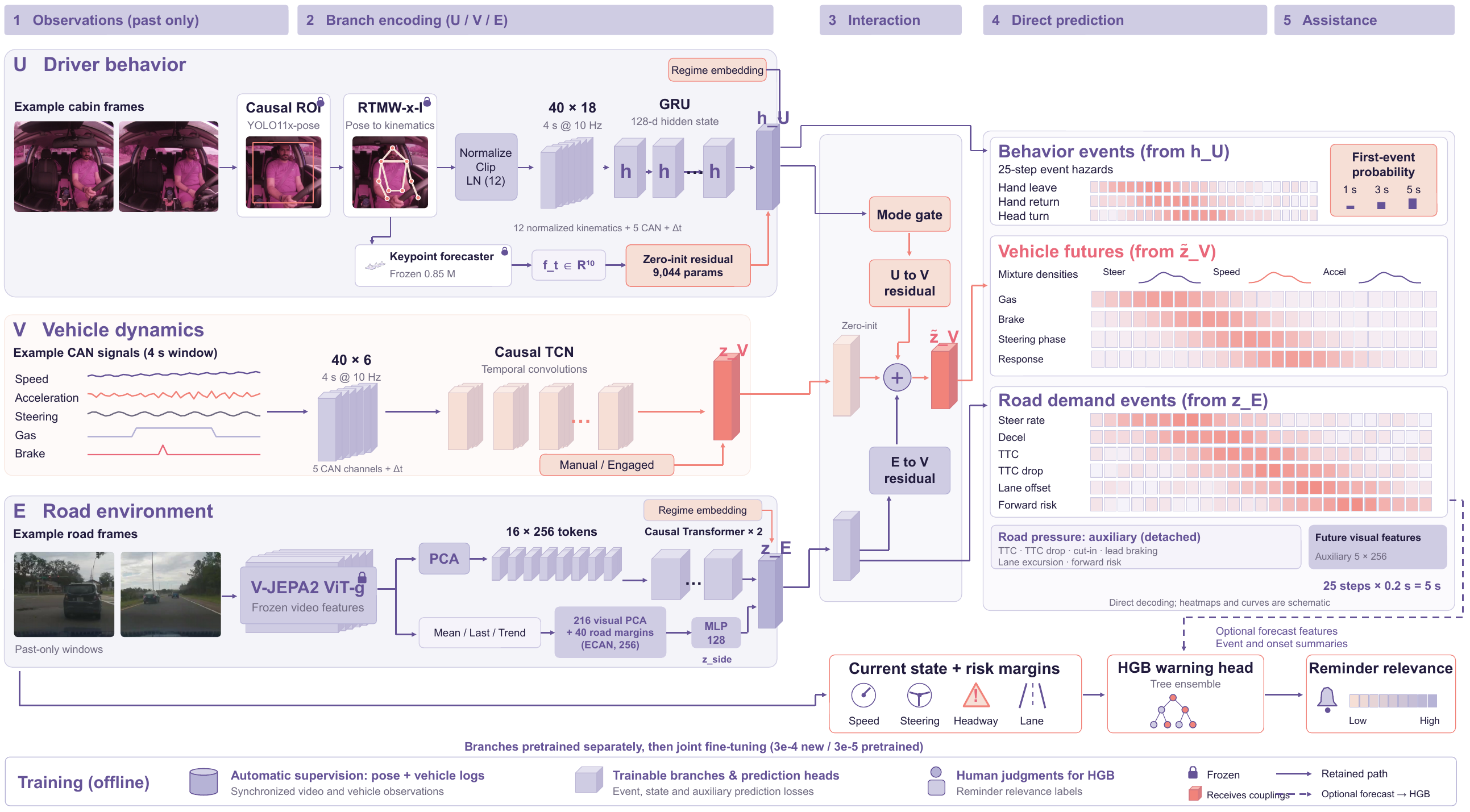}
    \caption{
        \textbf{Architecture of \model{}.}
        Driver, vehicle, and road histories feed joint forecasting; locks mark frozen modules, coral marks the vehicle state that receives the residual connections, and the dashed path marks forecast features evaluated offline as probe inputs.
    }
    \label{fig:architecture}
\end{figure}

At time $t$, let $\mathcal H_t$ denote the available cabin, vehicle, and road history, and let $m_t$ indicate manual or assistance-engaged driving (Figure~\ref{fig:architecture}).
\model{} jointly predicts structured state transitions and future road latents from the observed history, conditioned on the automation state $m_t$ rather than on a planned action.
All future hazards are decoded directly from the encoded history, without video reconstruction or latent rollout.
The model predicts 25 future steps at 0.2-second intervals.
Twelve primary binary chains, represented through transition hazards, cover two driver states, four vehicle states, and six road-demand proxies.
Driver states describe hands on the wheel and head turned.
Vehicle states describe gas, brake, steering phase (steering rate above 45\,deg/s with hysteresis), and response (braking, deceleration below $-0.8$\,m/s$^2$, or fast steering).
Road-demand proxies are based on steering rate, deceleration, time to collision (TTC), TTC drop, lane offset, and forward risk.
Continuous outputs describe steering angle, speed, and acceleration.
Auxiliary targets include six road-pressure chains, a hard-response chain, driver kinematics, and future road latents.

Event labels are derived from recorded signals and pose estimates; warning labels are collected separately through human annotation.
The primary metric is onset PR-AUC within three seconds, evaluated where the relevant transition can occur and future labels are available.
Target definitions and masks appear in Appendix~\ref{app:targets}.

\subsection{Modality-specific representations}
\label{sec:branches}

The three branches are pretrained separately on their assigned targets.
Learned automation-state embeddings condition each branch on $m_t$.

\textbf{Driver kinematics.}
A frozen detector and whole-body pose estimator extract driver keypoints within a region maintained from past observations.
Eleven kinematic channels describe head and wrist positions, head-turn deviation, and hand-to-rest distances.
Together with a validity bit, five CAN signals, and a time offset, they form each 18-dimensional input.
A GRU summarizes four seconds at 10\,Hz into $h_U$.
The driver branch therefore sees vehicle history even before any connection is added.

\textbf{Vehicle dynamics.}
A temporal convolutional network encodes four seconds of steering angle, gas and brake flags, speed, acceleration, and time offset into $z_V$.

\textbf{Road context.}
A frozen V-JEPA~2 ViT-g encoder \citep{assran2025} extracts road-video features, which PCA fitted on the training split projects to 256 dimensions.
A causal Transformer summarizes 16 clip tokens into $z_E$.
Each token covers a two-second clip and tokens end 0.5 seconds apart, so the 16-token bank spans about 9.5 seconds of video.
A side MLP encodes 216 visual components and 40 road-margin features into $z_{\mathrm{side}}$.
This configuration (ECAN) includes lead-vehicle, lane, and ego-motion quantities without driver-monitor outputs.
Its visual-only ablation uses 256 features.

\subsection{Automation-conditioned joint dynamics and directed connections}

\textbf{Driver forecast features.}
A trained keypoint forecaster observes dense motion over two seconds and sparse history extending to ten seconds.
Six state probabilities, three trajectory-mode probabilities, and a validity bit form $f_t\in\mathbb{R}^{10}$.
A residual map injects $f_t$ before branch connections:
\begin{equation}
\widetilde h_U
= h_U + r_F(\operatorname{sg}(f_t)).
\label{eq:forecastinput}
\end{equation}
$\operatorname{sg}$ stops gradients through the forecast features.
The forecaster is frozen and retained at inference.

\textbf{Directed residual connections (coupling).}
The vehicle state is the connection target because driver actions and road demands both surface as vehicle motion, whose chains have the most reliable labels.
We therefore direct their information into the vehicle state:
\begin{equation}
\widetilde z_V
= z_V
+ r_{E\rightarrow V}([z_E;z_{\mathrm{side}}])
+ \sigma(g_{m_t})\odot r_{U\rightarrow V}(\widetilde h_U).
\label{eq:coupling}
\end{equation}
Both connection maps are two-layer MLPs with 640 hidden units.
Here $g_{m_t}$ is a learned automation-state-specific gate, $\sigma$ is the sigmoid, and $\odot$ denotes elementwise multiplication.
The driver-to-vehicle residual and all driver prediction heads read the same $\widetilde h_U$, so forecast features influence both driver and vehicle predictions.
Zero-initialized final layers in $r_F$, $r_{E\rightarrow V}$, and $r_{U\rightarrow V}$ allow joint training to start from the pretrained predictions.
Zero initialization guarantees the starting function, not preservation after training.
Connection experiments omit the forecast-feature input to isolate the contribution of Eq.~\eqref{eq:coupling}.

Excluding frozen perception, the base joint forward path has 3.50M parameters, and the frozen keypoint forecaster has 0.85M.
Architecture details appear in Appendix~\ref{app:method}.

Each binary chain uses history-conditioned, time-varying first-order transitions.
Transition hazards provide first-event probabilities and state probabilities at multiple horizons from the same outputs, while availability masks exclude unobserved transitions in incomplete future sequences.
For chain $c$ and future step $k$, the model directly predicts on- and off-transition hazards $h^+_{c,k}$ and $h^-_{c,k}$ from $\mathcal H_t$ and $m_t$.
Given chain state $s_{c,k}$ and availability mask $a_{c,k}$, the loss is
\begin{align}
\ell_{c,k}=-a_{c,k}\big[&
(1-s_{c,k-1})\{s_{c,k}\log h^+_{c,k}
+(1-s_{c,k})\log(1-h^+_{c,k})\}
\nonumber\\
&+s_{c,k-1}\{(1-s_{c,k})\log h^-_{c,k}
+s_{c,k}\log(1-h^-_{c,k})\}\big].
\end{align}
For an inactive chain, the probability of its first onset $T$ within $K$ steps is
\begin{equation}
p_c(T\le K\mid s_{c,0}=0,\mathcal H_t,m_t)
=1-\prod_{k=1}^{K}(1-h^+_{c,k}).
\label{eq:onset}
\end{equation}
Off-transition events, such as a hand leaving the wheel, are handled analogously.
State probabilities start from $\pi_{c,0}=s_{c,0}$ and follow
\begin{equation}
\pi_{c,k}
=\pi_{c,k-1}(1-h^-_{c,k})
+(1-\pi_{c,k-1})h^+_{c,k}.
\end{equation}

All future hazards are decoded directly without feeding predictions back into a latent rollout.
Continuous vehicle signals use three-component mixture densities \citep{bishop1994}.
Joint training combines the masked prediction objectives with a future road-latent loss; auxiliary heads read branch states with gradients stopped.
Training settings, loss weights, and model selection appear in Appendix~\ref{app:training}.

AIDE results use our AIDE pose forecaster: a 5.77M-parameter part-token Transformer that applies the same anchored kinematic-forecasting design and BATON pose pretraining.
Its design and pretraining are detailed in Appendix~\ref{app:aide}.

\subsection{Dual-horizon forecasting and warning}
\label{sec:warninggen}

The joint model's five-second forecasts form the slow horizon; warnings form the fast horizon.
Warnings are issued by a lightweight histogram gradient boosting (HGB) probe over 77 current-state features (current states, road margins, current kinematics, dwell statistics, face geometry, automation state, and speed), trained on human labels.
Forecast outputs are logged and evaluated as candidate probe inputs in Section~\ref{sec:warning}, where probes that add predicted futures are compared with current-state probes using the same non-forecast inputs.
An openpilot-based baseline trained to reproduce logged attention warnings provides the offline comparison; its outputs are excluded from our predictors' inputs.

\label{sec:deployment}
Full road-model inference delays cabin processing on the shared GPU.
V-JEPA~2 takes 3.29\,s per clip, and the OpenBADAS risk pipeline \citep{goldshmidt2025badas} takes 3.3--3.5\,s.
We distill both into a ViT-S/16 student that takes 16 frames at $224\times224$ pixels and predicts a 1,408-dimensional road feature and a risk logit.
Road features feed the forecasting branch, while the risk output provides current context for the warning probe.
Downstream models and warning thresholds remain unchanged.
Let $v_T,v_S$ denote teacher and student features, $r_T,r_S$ their risk logits, and $P,Q$ the fixed token and visual-side projections.
Training combines standardized feature error, cosine distance, and risk losses:
\begin{equation}
\begin{aligned}
\mathcal L_{\mathrm{distill}}={}&\sum_{R\in\{I,P,Q\}}\lambda_R\big[\operatorname{MSE}_{\mathrm{std}}(Rv_S,Rv_T)+\alpha_R d_{\cos}(Rv_S,Rv_T)\big]\\
&+\lambda_b\operatorname{BCE}(\sigma(r_S),\sigma(r_T))+\lambda_r\|r_S-r_T\|_2^2.
\end{aligned}
\label{eq:distill}
\end{equation}
Here $I$ is the identity, $d_{\cos}(a,b)=1-\cos(a,b)$, and $\lambda_R,\alpha_R,\lambda_b,\lambda_r$ are nonnegative weights.
A comma four runs openpilot and the warning probe, while a chestnut eGPU (AMD Radeon RX 9060, 8\,GB) runs perception and forecasting.
GPU jobs are scheduled cooperatively, without preemption, by per-model priority.
If required inputs are stale, openpilot's driver monitor handles warnings.
The student supplies risk scores with a ten-second validity window, and the teacher risk model is disabled in the final configuration.
Student versions and deployment settings are detailed in Appendix~\ref{app:deployment}.

\section{Experiments and downstream evaluations}
\label{sec:results}
\subsection{Experimental setup}
\label{sec:setup}
We use the synchronized BATON benchmark subset \citep{wang2026}: 197.2 hours from 162 drivers, split 96/17/49 for training/validation/test, plus 11.71 hours from 60 additional drivers on which frozen models are scored.
Event labels come from pose estimates and recorded signals, warning labels from human annotation; label validation supports four driver-event targets (Appendix~\ref{app:targets}).
AIDE \citep{yang2023} uses the official split with five observed and five predicted poses, scored by All-MPJPE and PCK@0.05.
The offline openpilot-based baseline learns to reproduce logged attention warnings; the on-road comparison uses openpilot's driver-monitoring system.
Forecasting experiments use three seeds unless stated otherwise and AdamW (Appendix~\ref{app:training}).
Paired bootstrap intervals resample drivers for BATON and clips for AIDE.
For three-seed comparisons, \sig{} marks differences whose intervals exclude zero in the same direction in at least two seeds and whose absolute mean difference exceeds its across-seed SD.

\begin{table}[htbp]
\caption{Keypoint forecasting on AIDE (mean $\pm$ seed SD): All-MPJPE (pixels), PCK@0.05 (\%), and $h{=}5$ error on the 5\%, 20\%, and 50\% highest-motion clips; bold: best per block.}
\label{tab:aide}
\centering\scriptsize
\newsavebox{\cwAideBox}
\newcommand{\cwAideBody}{%
\rowcolor{cwBg}Model & All-MPJPE $\downarrow$ & PCK@0.05 $\uparrow$ & $h{=}5$, top 5\% $\downarrow$ & $h{=}5$, top 20\% $\downarrow$ & $h{=}5$, top 50\% $\downarrow$\\
\midrule
\multicolumn{6}{@{}l}{\textit{Published (Driver-WM paper)}}\\
ST-GCN & 110.98 & 60.97 & -- & -- & --\\
SiMLPe & 106.38 & 63.45 & -- & -- & --\\
MotionBERT & 73.51 & 78.01 & -- & -- & --\\
Driver-WM & 71.47 & 71.66 & -- & -- & --\\
Zero velocity & 52.89 & 85.95 & -- & -- & --\\
\addlinespace[2pt]\cmidrule(lr){1-6}\addlinespace[1pt]
\multicolumn{6}{@{}l}{\textit{Our protocol, anchored formulation (3 seeds unless stated)}}\\
Driver-WM, our reproduction (123.7M) & \gsd{130.86}{3.11} & \gsd{54.16}{.80} & \gsd{237.83}{3.11} & \gsd{169.42}{3.28} & \gsd{147.71}{3.68}\\
Driver-WM architecture, adapted (85.7M) & \gsd{50.65}{.29} & \gsd{86.24}{.13} & \gsd{213.63}{1.55} & \gsd{129.74}{.92} & \gsd{91.42}{.69}\\
SiMLPe, anchored (1.04M) & \gsd{50.26}{.07} & \gsd{86.47}{.16} & \gsd{214.48}{1.25} & \gsd{128.47}{.38} & \gsd{91.21}{.25}\\
ST-GCN, anchored (0.53M) & \gsd{49.22}{.18} & \gsd{86.86}{.02} & \gsd{215.26}{2.13} & \gsd{127.22}{.47} & \gsd{89.60}{.24}\\
MotionBERT, fine-tuned (42.5M) & \gsd{48.71}{.26} & \gsd{86.92}{.08} & \gsd{213.85}{.81} & \textbf{\gsd{127.09}{.13}} & \textbf{\gsd{88.94}{.51}}\\
\rowcolor{cwLavL}Ours (5.77M, 5 seeds) & \textbf{\gsd{48.60}{.12}} & \textbf{\gsd{87.02}{.06}} & \textbf{\gsd{212.34}{.45}} & \gsd{127.46}{.77} & \textbf{\gsd{88.94}{.32}}\\
\addlinespace[2pt]\cmidrule(lr){1-6}\addlinespace[1pt]
\multicolumn{6}{@{}l}{\textit{With BATON pose pretraining}}\\
Driver-WM architecture, adapted (85.7M) & \gsd{49.12}{.14} & \gsd{86.78}{.10} & \gsd{212.02}{1.04} & \gsd{128.54}{1.10} & \gsd{89.54}{.52}\\
ST-GCN, anchored (0.53M, 5 seeds) & \gsd{48.82}{.16} & \gsd{86.98}{.06} & \gsd{214.25}{.85} & \gsd{126.30}{.41} & \gsd{88.89}{.18}\\
\rowcolor{cwLavL}Ours (5.77M, 5 seeds) & \textbf{\gsd{48.05}{.06}} & \textbf{\gsd{87.13}{.08}} & \textbf{\gsd{210.51}{.90}} & \textbf{\gsd{125.64}{.85}} & \textbf{\gsd{87.87}{.38}}\\
}%
\begin{lrbox}{\cwAideBox}\setlength{\tabcolsep}{0pt}\begin{tabular}{lrrrrr}\cwAideBody\end{tabular}\end{lrbox}%
\setlength{\tabcolsep}{\dimexpr(\linewidth-\wd\cwAideBox)/12\relax}%
\begin{tabular}{lrrrrr}
\toprule
\cwAideBody
\bottomrule
\end{tabular}
\end{table}
\begin{table}[htbp]
\caption{Semantic recognition on AIDE: macro-F1 (\%) for driver behavior (DBR), emotion (DER), traffic context (TCR), and vehicle condition (VCR), and All-MPJPE of the same checkpoint.}
\label{tab:aidesem}
\centering\scriptsize
\newsavebox{\cwSemBox}
\newcommand{\cwSemBody}{%
\toprule
\rowcolor{cwBg}Model & DBR $\uparrow$ & DER $\uparrow$ & TCR $\uparrow$ & VCR $\uparrow$ & All-MPJPE $\downarrow$\\
\midrule
Driver-WM (published) & 68.07 & 72.61 & 90.15 & 68.34 & 71.47\\
\rowcolor{cwLavL}Ours, view-gate variant & \textbf{\gsd{71.13}{.32}} & \textbf{\gsd{74.67}{.36}} & \textbf{\gsd{90.29}{.79}} & \textbf{\gsd{70.30}{1.02}} & \textbf{\gsd{49.71}{.75}}\\
\bottomrule
}%
\begin{lrbox}{\cwSemBox}\setlength{\tabcolsep}{0pt}\begin{tabular}{lrrrrr}\cwSemBody\end{tabular}\end{lrbox}%
\setlength{\tabcolsep}{\dimexpr(\linewidth-\wd\cwSemBox)/12\relax}%
\begin{tabular}{lrrrrr}
\cwSemBody
\end{tabular}
\end{table}
\subsection{Task 1: In-cabin kinematic forecasting (AIDE)}
Table~\ref{tab:aide} compares our AIDE pose forecaster with the published baselines and with the same baselines re-instantiated in our protocol.
With BATON pretraining it reaches $48.05\pm0.06$ All-MPJPE, against Driver-WM's published 71.47 and the strongest published reference, zero velocity, at 52.89.
Two design choices explain the gain: first, the anchored formulation predicts displacements around the last observed pose rather than absolute coordinates. Zero velocity is already a strong predictor on AIDE, and anchoring lets the model spend its capacity on deviations from it.
An otherwise identical absolute-coordinate model is 3.87 pixels worse, and anchoring alone brings the published ST-GCN from 110.98 to 49.22.
Second, pretraining on 3.4 million naturalistic BATON pose frames supplies motion priors that the 1,884 AIDE training clips cannot; the paired gain of 0.55 pixels also transfers to the ST-GCN and Driver-WM architectures, which remain above ours with and without it.
The comparison also shows that the gain does not come from model size: the Driver-WM architecture adapted to the same tokens has fifteen times more parameters, and the 42.53M fine-tuned MotionBERT stays above our 5.77M forecaster.
On the high-motion subsets themselves, the pretrained forecaster has the lowest last-step error among all evaluated models at the 5\%, 20\%, and 50\% quantiles (Table~\ref{tab:aide}), and every \model{} variant is far below our reproduction of the published Driver-WM pipeline (Appendix~\ref{app:aide}).
Driver-WM's single published high-motion value (155.82 at $h{=}5$ on the top 10\%, 60 clips) is below our 159.92 on that subset, where sampling noise is large and anchoring is conservative for sudden motions.
The same recipe extends to semantic recognition: with a view-gate over exterior tokens, the variant is above Driver-WM on all four macro-F1 targets while keeping a 22-pixel lower All-MPJPE (Table~\ref{tab:aidesem}).
Exterior context matters little for keypoint geometry at this horizon: removing road features from a semantic variant changes All-MPJPE by only 0.36 pixels, and a road-to-driver connection on BATON is inconclusive (Appendices~\ref{app:aide} and~\ref{app:baton}).

\begin{figure}[htbp]
\centering
\includegraphics[width=0.96\linewidth]{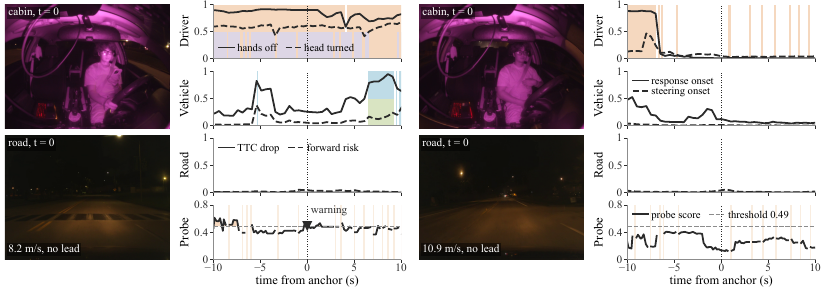}
\caption{\model{} on one study drive: sustained phone use (left) and a brief scratch during attentive driving (right). Curves: $P$(transition within 3\,s); bands: observed states.}
\label{fig:cases}
\end{figure}
\subsection{Task 2: Cross-domain dynamics anticipation (BATON)}
\label{sec:baton}
Figure~\ref{fig:cases} previews the joint model on one study drive: sustained phone use raises the driver branch's hands-off and head-turned probabilities while the vehicle and road branches stay quiet and the warning probe (Section~\ref{sec:warning}) crosses its threshold, whereas a brief scratch during attentive driving leaves all three branches and the probe low.

\textbf{Directed connections.}
The connections are directed into the vehicle state because driver actions and road demands both surface as vehicle motion, and that is where the gains appear (Table~\ref{tab:coupling}A).
Steering-onset PR-AUC rises by 0.030 in manual and 0.084 in assistance-engaged driving over independent branches, and manual braking improves as well.
Joint training without connections captures almost none of this, so the benefit comes from the connections rather than from shared optimization.
The engaged gain is plausibly larger because steering interventions under assistance are rare and hard to anticipate from vehicle history alone, but become predictable once the driver's hands and head enter the prediction.
Our all-to-all variant, which lets every branch rewrite every other with 43\% more connection parameters, is within 0.025 of the directed model on all vehicle and driver targets. Its clear gains are on the road-demand proxies (up to 0.167 on manual deceleration demand and 0.163 on engaged steering-rate demand), where the all-to-all road head can read the vehicle history from which these proxies are defined. 
We keep directed connections as the default because they leave the pretrained driver and road predictors intact, which keeps each branch's outputs interpretable and reusable, and because they use fewer parameters.

\textbf{Road inputs.}
Road video and structured road margins carry different information (Table~\ref{tab:coupling}B).
Replacing 40 visual components with the margins raises TTC-drop PR-AUC by 0.206 and 0.286, because time-to-collision is defined from the same radar quantities the margins expose; margins alone are even better on this and the other radar-defined targets.
Video tokens carry what margins cannot, forward risk and steering intent, where margins alone fall below ECAN.
The combined ECAN input is therefore the default, and both effects reproduce on 60 drivers never seen in training (Table~\ref{tab:coupling}C).

\textbf{Driver forecast features.}
Injecting the frozen keypoint forecaster's state probabilities helps the target that depends on where the driver is about to look: head-turn onset improves by about 0.026 in both automation states and by a similar amount on the additional drivers (Table~\ref{tab:coupling}D).
The hand-return gain does not transfer, consistent with hand-return having the lowest validated label precision of the four driver targets and with hands returning to the wheel being largely determined by the current state.
Six of the eight differences meet our criterion and none is negative; the remaining two are within noise, so forecast features never hurt a driver target and help most where the target depends on where the driver is about to look.
The model is also above a Driver-WM-style encoder adaptation on selected forecasting targets and above training from scratch; rollout decoders and current-state heuristics are compared in Appendix~\ref{app:baton}.

\begin{table}[htbp]
  \caption{Joint forecasting on BATON: onset PR-AUC within 3\,s (mean $\pm$ seed SD); peach cells with $\ast$ meet our criterion.}
  \label{tab:coupling}
  \centering\scriptsize

  \newsavebox{\cwBatonBox}
  \newcommand{\cwBatonBody}{%
    \toprule
    \multicolumn{6}{@{}l}{\rlap{\textbf{A. Connections (ECAN inputs, no forecast features); $\Delta$ = directed minus indep.\ branches / no links / all-to-all}}}\\[1pt]
    \rowcolor{cwBg} Driving & Event & $\Delta$ indep. & $\Delta$ no links & $\Delta$ all-to-all & Directed (default) \\
    \midrule 
    \multirow{3}{*}{Manual} 
      & Steering & \cellcolor{cwPeachL}+.030\sig & \cellcolor{cwPeachL}+.028\sig & \cellcolor{cwPeachL}$-$.007\sig & \cellcolor{cwLavL}\gsd{.774}{.002} \\
      & Gas      & \cellcolor{cwPeachL}+.020\sig & $-$.0000                      & $-$.011                         & \cellcolor{cwLavL}\gsd{.590}{.014} \\
      & Brake    & \cellcolor{cwPeachL}+.037\sig & \cellcolor{cwPeachL}+.039\sig & $-$.007                         & \cellcolor{cwLavL}\gsd{.381}{.003} \\
    \addlinespace[2pt]\cmidrule(lr){1-6}\addlinespace[1pt]
    \multirow{3}{*}{Engaged} 
      & Steering & \cellcolor{cwPeachL}+.084\sig & \cellcolor{cwPeachL}+.070\sig & +.002                           & \cellcolor{cwLavL}\gsd{.463}{.005} \\
      & Gas      & $-$.003                       & $-$.002                       & +.002                           & \cellcolor{cwLavL}\gsd{.033}{.003} \\
      & Brake    & +.003                         & +.003                         & +.003                           & \cellcolor{cwLavL}\gsd{.024}{.004} \\
    \midrule\addlinespace[3pt]

    \multicolumn{6}{@{}l}{\rlap{\textbf{B. Road inputs (benchmark test drivers) and the all-to-all variant on ECAN inputs}}}\\[1pt]
    \rowcolor{cwBg} Driving & Event & Visual-only & Sensor-only & All-to-all & ECAN, directed (default) \\
    \midrule
    \multirow{4}{*}{Manual} 
      & Steering onset        & \gsd{.774}{.002} & \gsd{.762}{.002} & \gsd{.781}{.001} & \cellcolor{cwLavL}\gsd{.774}{.002} \\
      & TTC-drop demand       & \gsd{.347}{.002} & \gsd{.610}{.003} & \gsd{.561}{.006} & \cellcolor{cwLavL}\gsd{.553}{.007} \\
      & Lead-braking pressure & \gsd{.172}{.005} & \gsd{.324}{.001} & \gsd{.242}{.004} & \cellcolor{cwLavL}\gsd{.235}{.003} \\
      & Forward-risk demand   & \gsd{.292}{.006} & \gsd{.205}{.002} & \gsd{.300}{.009} & \cellcolor{cwLavL}\gsd{.296}{.012} \\
    \addlinespace[2pt]\cmidrule(lr){1-6}\addlinespace[1pt]
    \multirow{4}{*}{Engaged} 
      & Steering onset        & \gsd{.461}{.003} & \gsd{.430}{.006} & \gsd{.461}{.006} & \cellcolor{cwLavL}\gsd{.463}{.005} \\
      & TTC-drop demand       & \gsd{.192}{.009} & \gsd{.556}{.015} & \gsd{.484}{.035} & \cellcolor{cwLavL}\gsd{.478}{.031} \\
      & Lead-braking pressure & \gsd{.125}{.006} & \gsd{.283}{.003} & \gsd{.174}{.022} & \cellcolor{cwLavL}\gsd{.175}{.017} \\
      & Forward-risk demand   & \gsd{.207}{.005} & \gsd{.154}{.002} & \gsd{.208}{.003} & \cellcolor{cwLavL}\gsd{.211}{.004} \\
    \midrule\addlinespace[3pt]

    \multicolumn{6}{@{}l}{\rlap{\textbf{C. Road inputs and the all-to-all variant, 60 additional drivers}}}\\[1pt]
    \rowcolor{cwBg} Driving & Event & Visual-only & Sensor-only & All-to-all & ECAN, directed (default) \\
    \midrule
    \multirow{3}{*}{Manual} 
      & Steering onset        & \gsd{.769}{.005} & \gsd{.752}{.005} & \gsd{.778}{.005} & \cellcolor{cwLavL}\gsd{.770}{.004} \\
      & TTC-drop demand       & \gsd{.250}{.018} & \gsd{.570}{.006} & \gsd{.496}{.006} & \cellcolor{cwLavL}\gsd{.459}{.009} \\
      & Lead-braking pressure & \gsd{.166}{.001} & \gsd{.372}{.003} & \gsd{.302}{.022} & \cellcolor{cwLavL}\gsd{.275}{.004} \\
    \addlinespace[2pt]\cmidrule(lr){1-6}\addlinespace[1pt]
    \multirow{3}{*}{Engaged} 
      & Steering onset        & \gsd{.314}{.009} & \gsd{.291}{.006} & \gsd{.321}{.016} & \cellcolor{cwLavL}\gsd{.319}{.009} \\
      & TTC-drop demand       & \gsd{.283}{.009} & \gsd{.494}{.005} & \gsd{.439}{.004} & \cellcolor{cwLavL}\gsd{.433}{.008} \\
      & Lead-braking pressure & \gsd{.114}{.003} & \gsd{.281}{.006} & \gsd{.157}{.002} & \cellcolor{cwLavL}\gsd{.156}{.003} \\
    \midrule\addlinespace[3pt] 

    \multicolumn{6}{@{}l}{\rlap{\textbf{D. Driver forecast features (default model: ECAN + frozen keypoint forecaster)}}}\\[1pt]
    \rowcolor{cwBg} Driving & Event & Default & $\Delta$ benchmark & $\Delta$ 60 add.\ drivers & \\
    \midrule
    \multirow{3}{*}{Manual} 
      & Hand-leave  & \cellcolor{cwLavL}\gsd{.436}{.010} & \gsd{+.0059}{.0004}                     & \cellcolor{cwPeachL}\textbf{\gsd{+.015}{.004}}\sig & \\
      & Hand-return & \cellcolor{cwLavL}\gsd{.826}{.007} & \cellcolor{cwPeachL}\textbf{\gsd{+.011}{.001}}\sig & \gsd{+.004}{.001}                                  & \\
      & Head-turn   & \cellcolor{cwLavL}\gsd{.474}{.002} & \cellcolor{cwPeachL}\textbf{\gsd{+.026}{.004}}\sig & \cellcolor{cwPeachL}\textbf{\gsd{+.030}{.004}}\sig & \\
    \addlinespace[2pt]\cmidrule(lr){1-6}\addlinespace[1pt]
    \multirow{1}{*}{Engaged} 
      & Head-turn   & \cellcolor{cwLavL}\gsd{.285}{.002} & \cellcolor{cwPeachL}\textbf{\gsd{+.027}{.005}}\sig & \cellcolor{cwPeachL}\textbf{\gsd{+.025}{.005}}\sig & \\
    \bottomrule
  }%
  \begin{lrbox}{\cwBatonBox}\setlength{\tabcolsep}{0pt}\begin{tabular}{llrrrr}\cwBatonBody\end{tabular}\end{lrbox}%
  \setlength{\tabcolsep}{\dimexpr(\linewidth-\wd\cwBatonBox)/12\relax}%
  \begin{tabular}{llrrrr}
    \cwBatonBody
  \end{tabular}
\end{table}

\subsection{Task 3: Real-time contextual warning probe}
\paragraph{Offline warning classification}
\label{sec:warning}
On 509 human-labeled anchors with 63 warranted warnings, the deployed current-state probe is above the openpilot-based baseline: AUROC 0.725 versus 0.563 in manual driving, where the difference meets our criterion, and 0.716 versus 0.637 pooled; in assistance-engaged driving the baseline is higher (0.746 versus 0.664).
The manual advantage comes from what the probe can see: road margins and dwell and face statistics let it separate a glance at a red light from the same glance while the gap ahead closes, whereas the baseline reproduces attention warnings that ignore the road.
The engaged reversal is plausibly the mirror image: openpilot's monitor is designed for engaged driving and its attention rule matches how raters judged warnings there, while the engaged estimate rests on only 32 warranted warnings, the smallest of the three evaluation strata.
The instantaneous warning decision thus favors low-variance current-state context, whereas the joint model's value lies in multi-second anticipation (TTC drop, steering onset; Section~\ref{sec:baton}).
Adding forecast features to the probe yields no consistent improvement in pooled AUROC ($-0.004\pm0.013$; Appendix~\ref{app:warning}). We therefore retain the current-state probe, which uses observed driver, vehicle, and road context together with recent-history summaries. The forecasting model runs alongside the warning probe, with forecasting accuracy and warning performance evaluated separately.
The probe is also better calibrated (expected calibration error 0.028 versus 0.408) and recalls more warranted warnings at the same warning load (Appendix~\ref{app:warning}).

\paragraph{On-vehicle runtime}
\label{sec:vehicle}
The final configuration updates warning scores at 5\,Hz with 135/177\,ms median/p95 driver-frame-to-score latency, while the forecasting backbone runs at 2.35\,Hz (target 2.5), pose at 9.38\,Hz, and the road student at 1.89\,Hz.
On four replay routes (two validation, two test) with the student's road features and risk substituted for the teacher's, the deployed student reproduces every teacher warning event (2/2, 6/6, 6/6, and 2/2 events), with 99th-percentile score differences of 0.007--0.036 and maximum differences of at most 0.081 (Appendix Table~\ref{tab:replay}).
Replaying the vehicle branch on the 13 study routes (5.0 hours of CAN) gives AUROC 0.907 for steering-phase onset within 3\,s, and the predicted probability rises from 0.19 at 5--10\,s before onset to 0.64 within 1\,s (Appendix~\ref{app:deployment}).

\begin{figure}[htbp]
\centering
\includegraphics[width=\linewidth]{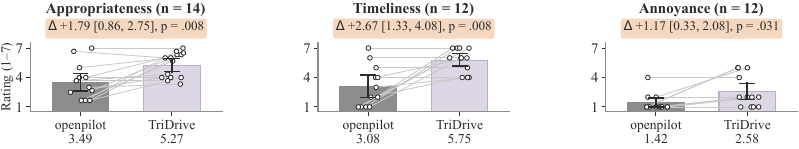}
\caption{Paired on-road ratings (1--7) with 95\% bootstrap intervals; grey lines join each participant's two ratings; $\Delta$ is \model{} minus openpilot (interval, Wilcoxon $p$).}
\label{fig:userstudy}
\end{figure}

\paragraph{Paired on-road user study}
\label{sec:userstudy}
Fourteen participants compared \model{} with openpilot's driver-monitoring system in their own vehicles, each driving the same route under both conditions.
The systems ran on two identical-looking comma four devices with system identity concealed from participants; condition order was alternated and documented in timestamped recordings of every drive and questionnaire session.
The 8.4 hours of dual-device recordings will be released with the paper, subject to consent and de-identification; protocol details and all participant scores are in Appendix~\ref{app:userstudy}.
Participants rated appropriateness (the primary outcome, averaging three items), timeliness, and annoyance on seven-point scales and selected an overall preference; we use participant-level paired bootstrap intervals (2,000 resamples) and two-sided Wilcoxon signed-rank tests.
Compared with openpilot, \model{} receives higher ratings for appropriateness by 1.79 points and timeliness by 2.67 points, alongside 1.17 points higher annoyance (Figure~\ref{fig:userstudy}).
Comments call \model{} ``quick and accurate'' at crossing pedestrians and stopping vehicles, describe openpilot's warnings as infrequent or late, and also name frequent or loud warnings.
The annoyance increase is the other side of the same design: a probe that warns on road context fires more often than one that warns only on sustained inattention, and comments cite frequency and loudness, not wrong situations; overall preference (7 \model{}, 4 openpilot, 3 no difference) is inconclusive.

\section{Discussion and conclusion}
\label{sec:discussion}
\textbf{Limitation.} The on-road study exercised only the fast horizon: the joint model ran on the vehicle, but its forecasts did not enter the warning decision, because adding them did not consistently raise offline warning AUROC and a current-state policy kept the warnings deterministic for participants. The deployment nevertheless establishes the system-level premise: the joint model, the distilled road encoders, and the probe run concurrently on the vehicle within the latency budget, and the logged forecasts are elevated at the anchors where warnings were presented (Appendix~\ref{app:deployment}). Forecast-conditioned warnings and action-conditioned rollouts remain future work

\model{} is a unified joint model of the driver, the vehicle, and the road, with driver monitoring as one downstream task. Participants rate the deployed warnings as more appropriate and timely, but also more annoying; the instantaneous warning decision favors low-variance current-state context, while the joint model's contribution is multi-second anticipation. Automatic labels miss some forms of inattention,  and the user study lacks a complete comparison of warning exposure.\label{lastmainpage}

\clearpage
\subsection*{AI use statement}
Generative AI tools were only used for translating and proofreading parts of the manuscript.
The authors are responsible for verifying the methods, numerical results, citations, and claims in the final manuscript.

\subsection*{Ethics statement}
The naturalistic recordings and the on-road study are covered by institutional review board approval, as reported for this research.
The participant study uses ordinary driving without induced distraction or secondary tasks.
Cabin recordings contain identifiable faces, and derived keypoints and features do not eliminate all privacy risks.
Video and annotation release is limited by consent and applicable access controls.
Participant identities are kept separately from analysis identifiers.
System identity was concealed from participants, and every drive and questionnaire session was video-recorded with consent.
Reported ratings concern warning experience; neither the offline evaluations nor the on-road study establishes crash reduction, clinical impairment, or improved vehicle control.

\subsection*{Reproducibility statement}
The appendices specify input tensors, targets, training schedules, model identities, evaluation populations, uncertainty estimates, deployment adaptations, and the fielded questionnaire.
Published baselines are distinguished from reimplementations and architecture adaptations.
Offline teacher results, deployed v5 measurements, and later student-fidelity results are reported separately.
The deployed warning path (current-state probe) and the logged forecasting backbone are identified separately, and the warning experiment on learned latent states is reported with its planned comparison.

To facilitate reproducibility, all survey questionnaires, source code, and key model checkpoints used in this work are provided in the supplementary material, with additional assets hosted at the public repository: \url{https://huggingface.co/HenryYHW/TriDrive}.
Furthermore, the supplementary material includes a selection of in-cabin experimental photos and recorded video clips, which are released with explicit participant consent, properly de-identified, and free of privacy concerns.
All associated code, full training pipelines, configuration manifests, and on-road user study data will be fully open-sourced in a public repository.

\bibliography{references}
\bibliographystyle{iclr2027_conference}
\clearpage
\appendix
\section{Model and data specification}
\label{app:method}

\subsection{Model variants and parameter accounting}

Because AIDE provides only observed poses in image coordinates while BATON provides synchronized pose, CAN-bus, and video histories; Table~\ref{tab:variants} distinguishes them so that geometry, 
semantics, event forecasting, and warning results are attributed correctly. 
Frozen perception models are excluded from the parameter totals below and remain part of the computational pipeline.

\begin{table}[h]
\caption{Model variants. Registered counts include modules that may not participate in the evaluated forward path.}
\label{tab:variants}
\centering\small
\begin{tabularx}{\linewidth}{L{3.0cm}rX}
\toprule
Configuration & Parameters & Role\\
\midrule
BATON, visual-only & 3,502,925 & Active forward-path count before forecast injection; road side vector is visual.\\
BATON, ECAN & 3,502,925 & Same active count; 40 visual side components replaced by road-margin features.\\
ECAN registered model & 3,649,913 & Includes an additional off-path pressure head.\\
Default registered joint model & 3,658,957 & Adds the 9,044-parameter forecast injector.\\
Compact pose forecaster & 850,627 & Trained separately, then frozen; retained at inference.\\
Default combined registry & 4,509,584 & Joint registry plus frozen forecaster, not 4.51M jointly trainable parameters.\\
AIDE pose forecaster & 5,771,449 & Part-token motion forecaster.\\ 
AIDE with semantic heads & 6,309,581 & Adds frozen visual tokens and simple semantic heads.\\
AIDE view-gate variant & 73,527,005 & Separate semantic attention heads and view gate.\\
\bottomrule
\end{tabularx}
\end{table}

The 3.50M count denotes the active base joint path before forecast-feature injection. The registered joint model includes the forecast injector and off-path heads, giving 3.66M registered parameters. Adding the separately trained 0.85M-parameter frozen forecaster gives a 4.51M combined registry. These counts exclude frozen perception and should not be interpreted as measurements of runtime memory.

The two residual connections contain 329,088 and 247,680 parameters for the environment-to-vehicle and driver-to-vehicle paths, respectively. Their total is 576,768, compared with 821,760 in the all-to-all fusion control. The residual maps are two-layer MLPs with 640 hidden units. Their final layers, together with the final layer of the 9,044-parameter forecast injector, are zero-initialized.

The active default count obtained by adding the injector and retained forecaster to the reported base forward path is 4,362,596. This arithmetic excludes the same off-path pressure head as the base forward count; it is not a measurement of inference memory or runtime. The per-module sums exclude 24 driver input-normalization parameters.
Parameter counts in this paper therefore state which quantity they refer to: the active forward path (3.50M), the registered joint model (3.66M), the combined registry with the frozen forecaster (4.51M), or the active default path (4.36M).

\textbf{Default forward path.}
The default model enables environment-to-vehicle and driver-to-vehicle residuals, while the optional road-to-driver connection is disabled. The forward order is GRU encoding, forecast-feature injection, optional road-to-driver connection, environment-to-vehicle connection, and driver-to-vehicle connection. Because the driver state is overwritten after forecast injection, the driver-to-vehicle residual reads $\widetilde h_U$, and all driver prediction heads read the same forecast-augmented state. At initialization, zero initialization gives $\widetilde h_U=h_U$ and $\widetilde z_V=z_V$; joint fine-tuning can subsequently change both states.

\subsection{Perception and input tensors}

YOLO11x-pose proposes person boxes at input size 384 and confidence threshold 0.3. Boxes covering less than 2\% of the image are ignored. RTMW-x-l estimates whole-body keypoints from the selected box with 10\% padding, maps them to HALPE-136, and masks keypoints below confidence 0.3. The road encoder is the frozen \texttt{facebook/vjepa2-vitg-fpc64-384} model. It receives 16 frames over each two-second clip. A training-fitted PCA reduces the pooled 1,408-dimensional feature to 256 dimensions. The semantic AIDE variants additionally use frozen 2,048-dimensional per-view features from Qwen3-VL-2B-Instruct.

The road token bank contains 16 tokens formed from two-second clips whose endpoints are spaced by 0.5 seconds. Thus, the bank covers approximately 9.5 seconds of past video at each anchor. The manual, engaged, and unknown automation state codes are mapped to learned embeddings before conditioning the corresponding branches.

\begin{table}[h]
\caption{BATON input tensors, omitting the batch dimension. Targets are not network inputs.}
\label{tab:inputs}
\centering\small
\begin{tabularx}{\linewidth}{L{2.1cm}L{2.0cm}X}
\toprule
Input & Shape & Contents\\
\midrule
Vehicle history & $40\times6$ & Five standardized vehicle signals and relative time, at 10\,Hz.\\
Driver history & $40\times18$ & Eleven kinematic channels, validity, five vehicle signals, relative time.\\
Road token bank & $16\times1\times256$ & Pooled clip features; separate presence flags and actual time lags.\\
Road side vector & $256$ & Visual PCA components, or 216 visual and 40 road-margin components in ECAN.\\
Automation state & scalar & Manual, engaged, or unknown.\\
Forecast input & $10$ & Six driver-state probabilities, three mode probabilities, and validity.\\
\bottomrule
\end{tabularx}
\end{table}

The 256-dimensional road side vector in Table~\ref{tab:inputs} is the input to the side encoder; $z_{\mathrm{side}}$ denotes its encoded representation. The environment's 40 nonvisual features consist of ten current road-margin quantities and 30 history summaries. The summaries include mean, slope, minimum, and maximum of lead distance, relative speed, lead speed, lane-center offset, and ego speed; mean lane width; lead and lane validity fractions; lead-target switch count; minimum TTC, headway, lateral margin, and lead acceleration; and approaching and excursion fractions. Standardization uses training-split statistics. Driver-monitor fields are excluded.

The driver association selects the person nearest a driver-side horizontal-position prior within 0.18 normalized image units.
The prior is estimated per recording from solo-occupant observations, with population and within-route fallbacks.
Running box medians are warmed up over 300 moving frames, expanded to 600, and then frozen.
Wrist rest positions and head-angle references use preceding valid observations, and kinematics are linearly interpolated from 5 to 10\,Hz.

\subsection{Targets, censoring, and label scope}
\label{app:targets}

Each anchor has 25 future steps at $t+0.2,\ldots,t+5.0$ seconds. Gas and brake targets use recorded pedal flags. Steering phase uses hysteresis on absolute steering rate, with on/off thresholds of 45/22.5 degrees per second. The response state is active when braking, acceleration below $-0.8$\,m/s$^2$, or lagged steering rate above 30 degrees per second is observed. A hard-response auxiliary state uses acceleration below $-2.5$\,m/s$^2$ or steering rate above 100 degrees per second. Each target step aggregates the two corresponding 10\,Hz samples.

Hands-on indicates a minimum hand-to-rest distance below 0.10 in crop-normalized units. Head-turn uses hysteresis on absolute head-turn deviation, with thresholds 0.1941 and 0.1026 fitted from training data. States reset after gaps longer than one second. Valid driver labels require sufficient confidence for the nose, both shoulders, and both wrists, a finite head-turn estimate, and an available reference.

The six primary demand chains represent high steering rate, deceleration, low TTC, falling TTC, lane offset, and forward risk. Their thresholds are respectively $\max(100\,\mathrm{deg/s},P_{98})$, $\min(-2\,\mathrm{m/s^2},P_2)$, TTC below four seconds, TTC derivative below $-1$\,s/s with TTC below eight seconds, lane offset above $\max(0.45\,\mathrm{m},P_{95})$, and a risk score above $\max(0.85,P_{95})$. These route quantiles are retrospective label definitions. They should not be confused with quantities known to an online predictor at the anchor.

Auxiliary road-pressure chains retain the TTC, TTC-drop, and forward-risk definitions and add cut-in, lead braking, and lane excursion. A cut-in is a new relevant lead after at least 0.5 seconds without one, or a range decrease exceeding five meters over a 0.1-second sample, with headway below 2.5 seconds; it is held for one second. Lead braking requires the same tracked lead to decelerate by more than 2\,m/s$^2$ over a backward one-second difference, with headway below four seconds. Lane excursion requires valid lane lines, width between 2.5 and 5.0 meters, and lateral margin below 0.9 meters. Missing lane observations are invalid rather than negative labels.

Futures are censored at route boundaries. Manual anchors are additionally censored when assistance engages; engaged anchors are not censored upon disengagement. An engaged anchor requires four seconds of engaged history without a pedal press, speed above 0.5\,m/s, and at least four seconds since engagement. These asymmetric rules define the evaluated populations and limit direct comparisons between automation states. Road-family validity uses motion and signal-presence rules; stationary observations are valid negatives where specified.

Human review of 33--34 events per class estimated label precision of 0.85, 0.82, 0.97, and 1.00 for manual hand-leave, hand-return, head-turn, and engaged head-turn, respectively. These small samples do not establish label recall or comprehensive behavioral coverage.

\subsection{Hazard interpretation and training}
\label{app:training}

With binary state $s_{c,k}$ and observability mask $a_{c,k}$, the transition likelihood is
\begin{align}
\ell_{c,k}=-a_{c,k}\big[&(1-s_{c,k-1})\{s_{c,k}\log h^+_{c,k}+(1-s_{c,k})\log(1-h^+_{c,k})\}\nonumber\\
&+s_{c,k-1}\{(1-s_{c,k})\log h^-_{c,k}+s_{c,k}\log(1-h^-_{c,k})\}\big].
\end{align}

The on-hazard is $P(s_k=1\mid s_{k-1}=0,\mathcal H_t,m_t)$, and the off-hazard is $P(s_k=0\mid s_{k-1}=1,\mathcal H_t,m_t)$. The model emits both for every future step. The observed previous state selects the appropriate likelihood term during training but is not supplied as a future network input. Eq.~\eqref{eq:onset} is the survival complement for first onset; for hand-leave, the analogous expression uses off-hazards and an initially active hands-on chain. The first-onset mass is
\[
q_k=h^+_k\prod_{j<k}(1-h^+_j).
\]
The marginal recurrence sums over possible state paths, while the onset score concerns the first transition. These quantities are not interchangeable.

\begin{table}[h]
\caption{Training schedules used in the experiments. Epoch counts are caps except where an executed pretraining duration is shown.}
\centering\small
\begin{tabular}{lrrrr}
\toprule
Stage & Learning rate & Weight decay & Batch & Epochs / patience\\
\midrule
Vehicle base & $3\times10^{-4}$ & .01 & 1,024 & 40 / 8\\
Driver base & $10^{-3}$ & $10^{-4}$ & 4,096 & 12 / 2\\
Road base & $3\times10^{-4}$ & .01 & 1,024 & 40 / 8\\
Joint, new/base & $3\times10^{-4}/3\times10^{-5}$ & .01 & 1,024 & 15 / 4\\
AIDE & $3\times10^{-4}$ & .05 & 32 & 200 / 40\\
BATON pose pretraining & $3\times10^{-4}$ & .05 & 256 & 6 executed\\
\bottomrule
\end{tabular}
\end{table}

All stages use AdamW. The joint stage draws 800,000 slice-balanced rows per epoch, uses three warm-up epochs followed by cosine decay, bf16 autocast, and gradient clipping at one. The vehicle objective combines three mixture-density losses and four transition losses, with steering phase weighted 0.5. The driver objective combines two transition losses and a detached kinematics loss weighted 0.5. The environment objective includes six demand chains, six detached pressure chains, and a future-latent cosine loss weighted 0.1. The detached hard-response head has weight one. Detachment prevents auxiliary losses from directly updating the encoder that supplies their features.

Checkpoint selection uses validation trunk-supervised chain NLL plus the weighted future-latent term. Continuous mixture, detached kinematics, pressure, and hard-response terms are excluded from that selection score. Hazards are clamped to $[10^{-5},1-10^{-5}]$, mixture scales use softplus plus $10^{-3}$, and Gaussian log-scales are clamped to $[-7,5]$. Training-set normalization is shared across assembled branches. The default forecaster remains frozen during joint fine-tuning.
\section{Additional BATON results}
\label{app:baton}
\subsection{Road information and connection controls}
Unless stated otherwise, the road-input and connection controls in this
subsection omit the forecast-feature injector. Their driver-to-vehicle
residual therefore receives $h_U$. The default model additionally injects
the frozen forecast features, so that both the driver heads and the
driver-to-vehicle residual receive $\widetilde h_U$.

Table~\ref{tab:roadinput} separates the ECAN input change from the directed connections. The two configurations have the same input width and network architecture; they differ in the information occupying 40 side-vector channels. Improved prediction of targets defined from radar, lane, and vehicle signals is evidence for the value of those histories. It is not evidence that visual features alone infer the same quantities.

\begin{table}[h]
\caption{ECAN minus visual-only environment input, before driver forecast injection. Entries are paired PR-AUC differences at 3\,s, mean $\pm$ seed SD. $\ast$ meets our criterion.}
\label{tab:roadinput}
\centering\small
\begin{tabular}{llrr}
\toprule
Family & Event & Manual & Engaged\\
\midrule
Demand & TTC & \sd{+.113}{.018}\sig & \sd{+.098}{.007}\sig\\
 & TTC drop & \sd{+.206}{.008}\sig & \sd{+.286}{.028}\sig\\
 & Lane offset & \sd{+.016}{.001}\sig & \sd{+.016}{.001}\sig\\
 & Deceleration & \sd{+.051}{.004}\sig & \sd{+.054}{.008}\sig\\
 & Steering rate & \sd{+.017}{.002}\sig & \sd{-.011}{.028}\\
Pressure & TTC & \sd{+.123}{.018}\sig & \sd{+.106}{.016}\sig\\
 & TTC drop & \sd{+.210}{.008}\sig & \sd{+.289}{.023}\sig\\
 & Lead braking & \sd{+.063}{.005}\sig & \sd{+.050}{.012}\sig\\
 & Lane excursion & \sd{+.033}{.006}\sig & \sd{+.014}{.005}\\
 & Cut-in & \sd{+.028}{.004} & \sd{+.097}{.001}\sig\\
 & Forward risk & \sd{+.006}{.007} & \sd{+.005}{.005}\\
Vehicle & Brake & \sd{+.017}{.004}\sig & \sd{-.003}{.002}\\
 & Gas & \sd{-.004}{.008} & \sd{-.002}{.003}\\
 & Steering phase & \sd{+.0002}{.0007} & \sd{+.001}{.008}\\
Driver & Hand-leave & \sd{+.0004}{.0007} & --\\
 & Hand-return & \sd{-.0003}{.0011} & --\\
 & Head-turn & \sd{+.0004}{.0002} & \sd{+.0007}{.0004}\\
\bottomrule
\end{tabular}
\end{table}
Table~\ref{tab:coupling}B places four road-input arms side by side on the targets where they differ most.
The sensor-only arm, which keeps the 40 margins and removes the video tokens, is above ECAN on TTC-drop demand (0.610/0.556 versus 0.553/0.478) and lead-braking pressure (0.324/0.283 versus 0.235/0.175).
It is below ECAN on forward-risk demand (0.205/0.154 versus 0.296/0.211) and manual steering onset (0.762 versus 0.774).
The margins therefore carry the radar-defined targets and the video tokens carry forward risk and steering; the combined ECAN input is retained because no single arm is above the others on every target.
All-to-all fusion on the ECAN input is within 0.01 of ECAN on every cell of Table~\ref{tab:coupling}B, consistent with Table~\ref{tab:coupling}.

The directed model's differences from all-to-all fusion are $-0.007$ for manual steering, $-0.025$ for manual head-turn, and $-0.167$ for manual deceleration demand; these meet our criterion.
Differences on driver endpoints against independent branches are inconclusive.
Against independent branches, manual gas NLL is worse by 0.030 nats despite its PR-AUC gain, and hand-return NLL by 0.014 nats, so improved ranking does not guarantee improved probabilistic fit.
The default model is above a label-free current-state heuristic, which scores each onset from the present hand-rest distance or head deviation, on all four validated driver endpoints.
Independent pretraining remains useful: on the visual-only configuration, training the fusion model from scratch decreases manual brake, gas, and steering PR-AUC by 0.074, 0.083, and 0.052 relative to the pretrained joint model, and manual hand-leave by 0.052.
Since architecture, schedule, and optimization history interact, this result supports the tested staged recipe rather than a universal principle that joint training is harmful.

\subsection{Additional-driver evaluation}
The additional set contains 60 route segments from 60 drivers absent from the benchmark.
It includes 116,801 manual anchors (6.49 hours) and 93,876 engaged anchors (5.22 hours).
The 42 vehicle identifiers do not overlap with benchmark vehicles.
Frozen candidates and comparators are scored on identical rows, without changing their thresholds or weights after inspecting that candidate's results.

The ECAN steering gains over independent branches are $0.046\pm0.004$ in manual driving and $0.075\pm0.011$ in engaged driving. Three of four designated vehicle comparisons reproduce the benchmark sign and verdict. For the ECAN input change, eight of 16 road comparisons reproduce. TTC-drop demand gains are $0.210\pm0.016$ and $0.149\pm0.004$, and lead-braking pressure gains are $0.109\pm0.004$ and $0.041\pm0.002$, respectively. The other comparisons are retained as inconclusive or nonreplicating, rather than combined into a general success claim.

The sensor-only and all-to-all arms were also scored on this set with their frozen benchmark checkpoints, with the comparison registered before scoring (Table~\ref{tab:coupling}C).
The sensor-only arm is again above ECAN on TTC-drop demand (0.570/0.494 versus 0.459/0.433) and lead-braking pressure (0.372/0.281 versus 0.275/0.156), on three of three seeds in both automation states, and below ECAN on manual steering onset (0.752 versus 0.770).
Forward-risk demand is not evaluable on this set because the risk labels are absent for its 60 routes.
All-to-all fusion is above ECAN on manual TTC-drop demand (0.496 versus 0.459) and manual steering onset (0.778 versus 0.770), and within 0.03 of ECAN on the other cells of Table~\ref{tab:coupling}C.

Driver forecast features improve head-turn PR-AUC on these drivers by $0.030\pm0.004$ and $0.025\pm0.005$, with all three seed intervals excluding zero (Table~\ref{tab:coupling}D). Head-turn NLL decreases by $0.0063\pm0.0003$ and $0.002\pm0.001$ nats. Manual hand-return changes by $0.004\pm0.001$, an inconclusive difference. 
\subsection{Keypoint forecasting diagnostics}
The compact forecaster predicts 11 upper-body joints at 0.5, 1, 2, and 3 seconds. Its dense-plus-sparse history model improves geometric error over zero velocity in all eight automation-state-by-horizon comparisons under our criterion. Improvements are largest during sustained motion. It remains worse than persistence on some short-horizon recovery subsets, so the overall result does not establish superiority for every motion phase.

The predicted head-turned state has AUROC 0.878, 0.839, and 0.808 at one, two, and three seconds in manual driving, versus 0.794, 0.728, and 0.687 for persistence.
Predicted hands-on state is worse than persistence at every tested horizon (0.749 versus 0.843 at one second).
These diagnostics are consistent with the most reproducible joint-model benefit appearing on head-turn events.
Road-demand and pressure predictions change by at most 0.005 PR-AUC after adding this driver input, an inconclusive difference.

\subsection{Road-state forecasting}
A secondary experiment predicts lead distance, relative speed, lead speed, and lane-center offset from eight seconds of bus history. A 149k-parameter GRU predicts a Gaussian residual over constant velocity. A matched variant also receives frozen road-video features. These predictors are separate diagnostic arms and do not replace the default model.

\begin{table}[h]
\caption{Road-state MAE at three seconds. Distance and lane offset are in meters; speeds are in meters per second. Learned values are mean $\pm$ SD across three seeds.}
\centering\small
\begin{tabular}{llrrrr}
\toprule
Driving & Target & Persistence & Constant velocity & Learned residual & + video\\
\midrule
Manual & Lead distance & 6.177 & 5.147 & \sd{4.735}{.034} & \sd{4.755}{.031}\\
 & Relative speed & 1.154 & 1.154 & \sd{.947}{.006} & \sd{.969}{.012}\\
 & Lead speed & 1.452 & 1.452 & \sd{1.303}{.012} & \sd{1.316}{.011}\\
 & Lane offset & .168 & .301 & \sd{.151}{.001} & \sd{.152}{.002}\\
Engaged & Lead distance & 4.892 & 5.328 & \sd{4.501}{.011} & \sd{4.549}{.081}\\
 & Relative speed & .807 & .807 & \sd{.813}{.012} & \sd{.823}{.016}\\
 & Lead speed & 1.017 & 1.017 & \sd{.943}{.011} & \sd{.960}{.012}\\
 & Lane offset & .093 & .190 & \sd{.087}{.000} & \sd{.088}{.000}\\
\bottomrule
\end{tabular}
\end{table}

The learned residual improves lead-distance error over constant velocity in both automation states. Its lead-braking onset PR-AUC is 0.227/0.224/0.224 across manual-driving seeds, versus 0.169 for the current-margin rule, and 0.232/0.227/0.222 versus 0.128 when engaged. Each paired interval excludes zero. Lane-excursion comparisons are inconclusive. Adding pooled video features does not improve the reported continuous predictions. 

\subsection{Architecture, robustness, and unsuccessful variants}
Most early architecture and input-perturbation experiments use the visual-only configuration. Their findings are not automatically assigned to the final ECAN model with forecast features. Table~\ref{tab:negative} summarizes the main outcomes, including unsuccessful approaches.

\begin{table}[h]
\caption{Additional comparisons and their scope. An inconclusive result does not establish equivalence or noninferiority.}
\label{tab:negative}
\centering\small
\begin{tabularx}{\linewidth}{L{3.0cm}X}
\toprule
Change & Evidence and interpretation\\
\midrule
Road-to-driver connection & Inconclusive against the two-connection model; omitted from the default.\\
Driver-WM-style encoder (21.2M Transformer) & The adapted trunk decreases manual hand-leave PR-AUC by .018 and steering by .009/.028 in manual/engaged driving. Not parameter-matched.\\
Matched driver encoders & A TCN decreases manual hand-leave by .020. A small Transformer improves head-turn ranking but worsens its hazard NLL.\\
Rollout (recurrent or stochastic) decoders & Some onset gains over fresh direct heads, but worse probabilistic or continuous scores and poorer results than pretrained direct heads.\\
Gaze-proxy chain & Head-turn improves by .023/.028, but adding its predictions gives no warning AUROC gain. Human gaze validation is incomplete.\\
Identity-invariance objectives & Adversarial heads and per-route normalization fail the joint identity-removal and prediction rule; driver endpoints worsen.\\
Joint overlap head & Does not satisfy the validation rule against the product of marginal forecasts.\\
Task-oriented latent objective & Changes no reported PR-AUC value by more than .0034 or NLL by .002 nats.\\
Phone/head-down proxies & Warning episode coverage improves at 20 alerts/hour but decreases at 40; not adopted.\\
Cabin latent features & High driver-identification accuracy (.91--.94) without supported driver-onset gains.\\
Spatial road tokens & No improved endpoint; manual steering-rate demand decreases by .019.\\
\bottomrule
\end{tabularx}
\end{table}

On the visual-only model, dropping 20\%/50\% of driver history decreases manual hand-leave PR-AUC by 0.076/0.110. Removing road tokens decreases steering PR-AUC by 0.014/0.051 in manual/engaged driving and manual TTC-drop pressure by 0.169. Noise of 0.5\,m/s in speed and two degrees in steering angle decreases engaged steering PR-AUC by $0.081\pm0.010$. Masking recent driver history is more harmful than masking the oldest two seconds. Insensitivity to a structurally disconnected modality is an architectural property, not learned robustness.

A driver-identity probe reaches top-1 accuracy 0.61/0.73 in manual/engaged driving, versus majority rates 0.25/0.14.
These results expose retained recording-context information.
These results do not prove that every prediction uses an identity shortcut, but they rule out claiming driver-invariant representations.
On 56 labeled phone episodes, the warning probe alerts on 21\% at 20 alerts/hour and 38\% at 40; lateral head-turn proxies miss many downward and in-lap phone behaviors.

\section{AIDE formulation, transfer, and baseline comparisons}
\label{app:aide}
\subsection{Protocol and output formulation}
AIDE clips contain 45 frames at 15\,Hz. Ten uniformly sampled frames define five observed and five future poses; the last predicted sample is not a five-second horizon. The official split has 1,884 training, 405 validation, and 609 test clips and is not subject-disjoint. High-motion evaluation uses the 60 clips with the greatest future displacement. MPJPE averages Euclidean joint errors in pixels. PCK@0.05 uses $0.05\max(W,H)=96$ pixels. Keypoints are estimates from a pose pipeline, including interpolated missing joints, rather than motion-capture ground truth.

Part tokens cover body joints 0--25, face 26--93, left hand 94--114, and right hand 115--135. The hands share a tokenizer and use side embeddings. Each joint contributes displacement relative to the last observed pose divided by 200, normalized position, and confidence. The Transformer has width 256, four layers, four heads, and feed-forward width 1,024. Its prediction can be written as
\begin{equation}
\widehat x_k=x_0+g_k\odot\left(\sum_{a=1}^{3}\alpha_{k,a}A_a-x_0\right)+200\mu_k,
\qquad \alpha_k=\operatorname{softmax}(u_k),
\end{equation}
where $x_0$ and $x_{-1}$ are the last and penultimate observed poses, $g_k$ is the horizon gate, $u_k$ are anchor-mixture logits, and $\mu_k$ is the predicted residual. The anchors are $A_1=x_0$, $A_2$ is the mean observed pose, and $A_3=x_0-0.1(x_0-x_{-1})$. The reported initialization reproduces $x_0$ within $10^{-6}$ pixels. This identity requires both the residual and effective anchor-mixture displacement to vanish.

The objective combines Euclidean pixel error, isotropic Laplace NLL with weight 0.1, bone-length regularization with weight 0.05, and temporal jerk with weight 0.01. Augmentation samples frame gaps of four or five, retains the protocol grid with probability 0.5, adds confidence-dependent keypoint noise, drops joints with probability 0.1, translates by up to 20 pixels, and scales within $[0.95,1.05]$. Horizontal flips are not used. Checkpoints use validation All-MPJPE of exponential-moving-average weights with decay 0.999.

\subsection{Published references and matched formulation}
\begin{table}[h]
\caption{Additional AIDE geometry references. HM is horizon-averaged high-motion MPJPE.}
\centering\small
\begin{tabular}{lrrr}
\toprule
Published method & All-MPJPE & HM-MPJPE & PCK@0.05\\
\midrule
ST-GCN & 110.98 & 158.10 & 60.97\\
SiMLPe & 106.38 & 156.29 & 63.45\\
MotionBERT & 73.51 & 141.53 & 78.01\\
Offline encoder--decoder & 75.87 & 144.05 & 77.17\\
Static pooling & 68.50 & 134.56 & 72.55\\
Single stream & 90.87 & 147.44 & 59.63\\
Late fusion & 82.59 & 143.31 & 65.07\\
Cross-attention only & 80.14 & 142.41 & 66.43\\
Driver-WM, 3 seeds & \sd{71.66}{.30} & \sd{138.34}{1.20} & \sd{71.40}{.30}\\
\bottomrule
\end{tabular}
\end{table}

\begin{table}[h]
\caption{AIDE geometry.
MPJPE is in pixels (lower is better); PCK is a percentage.
Published values are unpaired references from Driver-WM.
Our learned rows report mean $\pm$ seed SD.
HM denotes the 60 highest-motion clips; $h=5$ is the last predicted sample.
Ours is our AIDE pose forecaster (5.77M parameters), which applies \model{}'s anchored kinematic-forecasting design.}
\label{tab:aidefull}
\centering\small\setlength{\tabcolsep}{4pt}
\begin{tabular}{lrrrr}
\toprule
Model & All-MPJPE & HM-MPJPE & PCK & HM $h=5$\\
\midrule
Zero velocity (published) & 52.89 & 139.19 & 85.95 & 178.62\\
Driver-WM (published) & 71.47 & 138.03 & 71.66 & 155.82\\
Driver-WM, our reproduction (123.7M, 3) & \sd{130.86}{3.11} & \sd{176.33}{3.29} & \sd{54.16}{.80} & \sd{194.34}{2.65}\\
\midrule
Zero velocity (ours) & 52.93 & 139.23 & 85.95 & 178.67\\
ST-GCN, anchored (3 seeds) & \sd{49.22}{.18} & \sd{130.15}{.52} & \sd{86.86}{.02} & \sd{162.97}{1.05}\\
MotionBERT, fine-tuned (3) & \sd{48.71}{.26} & \sd{128.84}{.47} & \sd{86.92}{.08} & \sd{158.38}{.79}\\
Ours (3 seeds) & \sd{48.63}{.15} & \sd{131.64}{.76} & \sd{87.04}{.08} & \sd{163.37}{.82}\\
Ours (5 seeds) & \sd{48.60}{.12} & \sd{131.57}{.57} & \sd{87.02}{.06} & \sd{163.37}{.63}\\
ST-GCN, anchored, + BATON (5) & \sd{48.82}{.16} & \sd{129.91}{.48} & \sd{86.98}{.06} & --\\
Driver-WM arch., adapted, + BATON (3) & \sd{49.12}{.14} & \sd{130.00}{.24} & \sd{86.78}{.10} & \sd{163.27}{.63}\\
Ours + BATON pretraining (5) & \sd{48.05}{.06} & \sd{128.82}{.39} & \sd{87.13}{.08} & \sd{159.92}{.92}\\
\bottomrule
\end{tabular}
\end{table}

Our anchored forecaster reaches 48.63 across three seeds. An otherwise matched absolute-coordinate arm reaches $52.50\pm0.42$, a paired increase of $3.87\pm0.46$ with each seed interval excluding zero.

Reinstantiating generic encoders under the same anchored formulation gives $49.22\pm0.18$ for ST-GCN and $50.26\pm0.07$ for SiMLPe.
An adapted Driver-WM architecture operating on the same keypoint part tokens reaches $50.65\pm0.29$ with 85.72M parameters, compared with 5.77M for our AIDE pose forecaster.
Its paired difference is $2.02\pm0.45$ pixels; the seed-averaged clip-bootstrap interval is $[1.53,2.52]$.
This is an architecture adaptation, not the original Driver-WM input pipeline.
Our reproduction of the published Driver-WM pipeline (absolute sigmoid-normalized coordinates with an MSE loss, 123.7M parameters) reaches $130.86\pm3.11$ All-MPJPE and $194.34\pm2.65$ HM $h{=}5$ over three seeds, against the published 71.47 and 155.82 (Table~\ref{tab:aidefull}).
Every \model{} variant is below this reproduction on every metric and on every high-motion quantile, by 25.6 to 77.6 pixels at $h{=}5$ (paired, three of three seeds).
The published values could not be reproduced from the released description, so they are quoted as reported and compared without pairing.

\subsection{Naturalistic pose pretraining}
The BATON pose corpus uses 440 training routes from 96 benchmark training drivers. It contains 3,410,103 frames at 10\,Hz, with 68.7\% valid. Ten-sample windows at 0.3-second spacing produce 642,179 valid windows. Of these, 18,155 windows from 22 routes are reserved for pretraining checkpoint selection, leaving 624,024 training windows. A per-window similarity maps shoulder width to the AIDE training median of 309 pixels and shoulder center to $(373,762)$. The missing head-top point is approximated from nose and neck; confidences are mapped to the training distribution. This normalization is used for offline pretraining, not to assert a streaming pose transform.

Pretraining runs for six epochs; the reported selected checkpoint is epoch four. Fine-tuning transfers the tokenizer, encoder blocks, and final normalization, but not the horizon output head. Initial transfer of every weight is inconclusive ($-0.31\pm0.02$ pixels). Trunk-only transfer was first explored and then replicated with an independent pretraining seed, giving $-0.62\pm0.17$ on the three matched seeds and $-0.55\pm0.15$ across five. The five-seed All-MPJPE is $48.05\pm0.06$. The pooled paired interval is $[-0.94,-0.18]$. Alternative pretraining schedules do not establish a further improvement.

The same corpus also benefits the adapted Driver-WM architecture and ST-GCN.
The adapted Driver-WM gains $1.53\pm0.42$ pixels and remains $1.11\pm0.13$ above our pretrained pose forecaster on three matched seeds.
Under a common converged pretraining recipe, our pose forecaster is $0.78\pm0.15$ below ST-GCN over five seeds.

\subsection{Released MotionBERT weights}
The official DSTformer uses width 512, depth five, and eight heads. Its released weights are loaded with 258 of 260 tensors unchanged, omitting the 3-D lifting head. A joint map expands spatial embeddings to the 136-keypoint layout. Inputs contain five observations followed by five copies of the last observed pose; no actual future pose is supplied. All 42.53M parameters are fine-tuned under the AIDE protocol, with either anchored or absolute-coordinate output.

\begin{table}[h]
\caption{MotionBERT controls on AIDE, three seeds. The post-hoc no-EMA row addresses an optimization confound in the absolute-coordinate comparison.}
\centering\small
\begin{tabular}{lrr}
\toprule
Configuration & All-MPJPE & Status\\
\midrule
Released weights, anchored output & \sd{48.71}{.26} & Evaluated\\
Released weights, absolute output & \sd{53.18}{.37} & Evaluated\\
Random initialization, anchored & \sd{48.70}{.19} & Evaluated\\
Released weights, absolute, no EMA & \sd{49.31}{.22} & Post hoc\\
Ours (pose forecaster), no pretraining & \sd{48.63}{.15} & Reference\\
Ours (pose forecaster) + BATON, matched seeds & \sd{48.01}{.03} & Reference\\
\bottomrule
\end{tabular}
\end{table}

The anchored fine-tune is indistinguishable from its random-initialized counterpart under the tested schedule. Removing EMA improves the absolute-output arm by $3.87\pm0.47$ pixels, showing that much of its original deficit comes from averaging lag under the training cap. The remaining paired difference from the anchored arm is $0.60\pm0.04$. Thus, the initial 4.47-pixel gap should not be attributed solely to anchoring. This is a forecasting adaptation of released weights, not MotionBERT evaluated on its original 3-D lifting task. The same control on our forecaster leaves the anchoring gap at $4.30\pm0.18$ pixels without EMA (three of three seeds), so this confound is specific to MotionBERT.

The paired difference between anchored MotionBERT and our pretrained pose forecaster is $0.71\pm0.22$ pixels across the three matched seeds.
Two seed intervals exclude zero.
The five-seed value 48.05 is not substituted into this three-seed paired comparison.

\subsection{High-motion and unadopted variants}
The five-seed pretrained model's final-sample high-motion MPJPE is 159.92, above Driver-WM's published 155.82. The official high-motion subset contains only 60 clips and has substantial sampling uncertainty. This does not convert a worse point estimate into a win. Published hand-joint error on that subset is also lower than ours. Loss reweighting and learned residual-gain variants do not meet their stated improvement thresholds.


\subsection{High-motion quantile sweep}
The official high-motion subset is the 10\% of test clips with the largest future displacement (60 clips).
Tables~\ref{tab:hmsweeph5} and~\ref{tab:hmsweepavg} sweep this threshold from the top 5\% to the full set for every model in our protocol, using the published motion score; cells are seed means $\pm$ SD (number of seeds in parentheses where it differs from three).
Our pretrained forecaster has the lowest error at every quantile among all evaluated models, for both the last predicted pose and the horizon average, and every \model{} variant is 25--78 pixels below our reproduction of the published Driver-WM pipeline at $h{=}5$ (paired, three of three seeds at every quantile).
The published Driver-WM reports a single high-motion value at $h{=}5$ (155.82 on the top 10\%), which is below ours (159.92); the gap to our reproduction and to the adapted architecture shows that this number is not matched by either port of the published method.

\begin{table}[h]
\caption{MPJPE (pixels) of the last predicted pose ($h{=}5$) on the top-$q$ high-motion subsets of the AIDE test split (number of clips in parentheses).}
\label{tab:hmsweeph5}
\centering\scriptsize\setlength{\tabcolsep}{2.5pt}
\begin{tabular}{lrrrrrr}
\toprule
Model & 5\% (30) & 10\% (60) & 20\% (121) & 30\% (182) & 50\% (304) & all (609)\\
\midrule
Zero velocity & 230.28 & 178.67 & 140.66 & 120.46 & 97.28 & 63.74\\
Ours (5) & \sd{212.34}{.45} & \sd{163.37}{.63} & \sd{127.46}{.77} & \sd{109.85}{.49} & \sd{88.94}{.32} & \sd{57.64}{.18}\\
Ours + BATON pretraining (5) & \sd{210.51}{.90} & \sd{159.92}{.92} & \sd{125.64}{.85} & \sd{108.53}{.57} & \sd{87.87}{.38} & \sd{57.02}{.18}\\
Ours, view-gate semantic variant & \sd{211.05}{1.93} & \sd{161.93}{.84} & \sd{126.92}{.16} & \sd{109.64}{.30} & \sd{88.69}{.19} & \sd{57.47}{.14}\\
ST-GCN, anchored & \sd{215.26}{2.13} & \sd{162.97}{1.05} & \sd{127.22}{.47} & \sd{110.29}{.43} & \sd{89.60}{.24} & \sd{58.47}{.21}\\
SiMLPe, anchored & \sd{214.48}{1.25} & \sd{162.86}{.93} & \sd{128.47}{.38} & \sd{111.46}{.19} & \sd{91.21}{.25} & \sd{59.48}{.14}\\
Driver-WM architecture, adapted & \sd{213.63}{1.55} & \sd{162.83}{1.13} & \sd{129.74}{.92} & \sd{112.44}{.85} & \sd{91.42}{.69} & \sd{60.04}{.40}\\
ST-GCN, anchored + BATON (5) & \sd{214.25}{.85} & \sd{162.51}{.32} & \sd{126.30}{.41} & \sd{109.27}{.35} & \sd{88.89}{.18} & \sd{58.02}{.14}\\
Driver-WM architecture + BATON & \sd{212.02}{1.04} & \sd{163.27}{.64} & \sd{128.54}{1.10} & \sd{110.97}{.63} & \sd{89.54}{.52} & \sd{58.63}{.35}\\
Driver-WM, our reproduction & \sd{237.83}{3.11} & \sd{194.34}{2.65} & \sd{169.42}{3.28} & \sd{158.09}{3.48} & \sd{147.71}{3.68} & \sd{134.50}{3.46}\\
Driver-WM (published, single run) & -- & 155.82 & -- & -- & -- & --\\
\bottomrule
\end{tabular}
\end{table}
\begin{table}[h]
\caption{Horizon-averaged MPJPE (pixels) on the same top-$q$ high-motion subsets.}
\label{tab:hmsweepavg}
\centering\scriptsize\setlength{\tabcolsep}{2.5pt}
\begin{tabular}{lrrrrrr}
\toprule
Model & 5\% (30) & 10\% (60) & 20\% (121) & 30\% (182) & 50\% (304) & all (609)\\
\midrule
Zero velocity & 175.66 & 139.23 & 111.54 & 97.16 & 79.36 & 52.93\\
Ours (5) & \sd{166.13}{.81} & \sd{131.57}{.57} & \sd{103.97}{.59} & \sd{91.00}{.39} & \sd{74.01}{.27} & \sd{48.60}{.12}\\
Ours + BATON pretraining (5) & \sd{163.85}{.88} & \sd{128.82}{.39} & \sd{102.48}{.38} & \sd{89.81}{.26} & \sd{73.10}{.14} & \sd{48.05}{.06}\\
Ours, view-gate semantic variant & \sd{164.92}{1.49} & \sd{130.28}{.86} & \sd{103.40}{.49} & \sd{90.68}{.50} & \sd{73.69}{.30} & \sd{48.36}{.17}\\
ST-GCN, anchored & \sd{166.10}{.90} & \sd{130.15}{.52} & \sd{103.34}{.23} & \sd{91.01}{.26} & \sd{74.38}{.22} & \sd{49.22}{.18}\\
SiMLPe, anchored & \sd{166.36}{.33} & \sd{131.22}{.27} & \sd{104.91}{.31} & \sd{92.32}{.28} & \sd{76.04}{.20} & \sd{50.26}{.07}\\
Driver-WM architecture, adapted & \sd{168.27}{.99} & \sd{132.92}{.87} & \sd{106.44}{.65} & \sd{93.39}{.63} & \sd{76.24}{.54} & \sd{50.65}{.29}\\
ST-GCN, anchored + BATON (5) & \sd{165.29}{.55} & \sd{129.91}{.49} & \sd{102.84}{.35} & \sd{90.28}{.30} & \sd{73.80}{.27} & \sd{48.82}{.16}\\
Driver-WM architecture + BATON & \sd{164.37}{.57} & \sd{129.99}{.24} & \sd{104.26}{.49} & \sd{91.35}{.17} & \sd{74.08}{.24} & \sd{49.12}{.14}\\
Driver-WM, our reproduction & \sd{206.85}{3.91} & \sd{176.33}{3.29} & \sd{157.20}{3.29} & \sd{149.06}{3.29} & \sd{141.44}{3.31} & \sd{130.86}{3.11}\\
Driver-WM (published, single run) & -- & 138.03 & -- & -- & -- & 71.47\\
\bottomrule
\end{tabular}
\end{table}
\subsection{Semantic recognition and diagnostic interventions}
Semantic recognition uses the observed window and is distinct from future-state forecasting. The view-gate semantic model contains 73.53M parameters. Its checkpoint selected by validation mean macro-F1 differs from its geometry-selected checkpoint. Table~\ref{tab:semantics} reports each with its own geometry. The gains over the published Driver-WM main row are numerical comparisons; no paired test is possible without its predictions.

\begin{table}[h]
\caption{AIDE semantic macro-F1 (\%). DBR: driver behavior; DER: driver emotion; TCR: traffic context; VCR: vehicle condition.}
\label{tab:semantics}
\centering\small
\begin{tabular}{lrrrrr}
\toprule
Model / checkpoint & DBR & DER & TCR & VCR & All-MPJPE\\
\midrule
Driver-WM, published single run & 68.07 & 72.61 & 90.15 & 68.34 & 71.47\\
View-gate, geometry-selected & 67.56 & 73.06 & 91.39 & 69.94 & 48.64\\
View-gate, F1-selected & 71.13 & 74.67 & 90.29 & 70.30 & 49.71\\
Three-seed logit ensemble & 73.40 & 77.29 & 91.16 & 72.33 & 48.82\\
\bottomrule
\end{tabular}
\end{table}

Exterior-feature interventions change predictions, but sensitivity to an input is not evidence of causal identification. 
Removing exterior context from the 6.31M-parameter variant with simple semantic heads (Table~\ref{tab:variants}) changes predicted coordinates by 5.78 pixels while increasing All-MPJPE by only 0.36. 
Finally, ranking the 60 highest-motion clips gives predicted-motion AUROC 0.851, versus 0.781 for observed motion. 
This target is future motion magnitude, not independently labeled danger, so it is reported as a motion-ranking diagnostic rather than safety-risk detection.

\section{Warning probes and offline diagnostics}
\label{app:warning}
\subsection{Features, training, and evaluation populations}
The warning classifier is trained on 979 consensus-labeled training anchors, including 182 warranted warnings from 91 drivers. Candidate learners include logistic regression, histogram gradient boosting, and a small MLP, selected by driver-grouped cross-validation. The prespecified feature/learner comparison and the later dwell-and-face extension are distinct analyses. The latter is exploratory.

Observable current-state features include anchor states, road margins and validity, current kinematics, visibility, automation state, and speed.
Eleven dwell features summarize hands-off and head-away shares, time since recovery, maximum deviation, transition count, and valid history.
Sixteen face features summarize current landmark geometry, face validity, gaze-proxy shares, yaw deviation, eye openness, blink rate, and valid history.
These proxies are not validated eye-tracking measurements.
Together these 77 features are the inputs of the current-state probe deployed on the vehicle (Section~\ref{sec:deployment}).

The predicted block includes driver marginals, future kinematics, event-onset scores, vehicle-response marginals, road-pressure predictions, and hard-response predictions. Models using this block depend on the forecasting checkpoint; the current-state classifier does not.

The histogram gradient-boosting learner uses maximum depth three, L2 regularization one, learning rate 0.05 or 0.1, and 100 or 200 iterations. Monotonicity constraints encode selected feature directions; other features remain unconstrained. Five-fold GroupKFold uses driver identifiers. Any temperature calibration is fitted on clean validation labels only. The openpilot-based baseline learns the logged attention states of openpilot \citep{commaai2026openpilot}; it is a learned forecast of those alerts, not the actual deployed alert policy replayed on the same frames.

The main gold discrimination set has 509 anchors: 327 manual with 31 positives, and 182 engaged with 32 positives. Some operating-point, calibration, and oracle diagnostics use a 512-row anchored layout with the same 63 positives, adding two manual and one engaged negative. Those diagnostics are labeled by their actual population. This difference must not be silently treated as identical rows in paired analysis.

\subsection{Discrimination, out-of-fold features, and oracle diagnostics}
The deployed current-state probe reaches AUROC 0.716/0.725/0.664 in pooled/manual/engaged evaluations; its intervals are paired driver-bootstrap intervals of a single model, since the probe does not depend on the forecasting seed.

Forecast-dependent training features are also generated out of fold: each training driver is scored by branches trained without that fold's drivers.
The out-of-fold current-plus-future probe reaches AUROC 0.668/0.695/0.614, compared with 0.691/0.667/0.664 for the base current-state probe trained on the same out-of-fold layout without dwell and face features.
All three stratum differences are inconclusive.
Adding predicted features to the dwell-and-face probe changes pooled AUROC by $-0.004\pm0.013$ across seeds.

A prespecified experiment tests a warning probe on learned latent states: the joint model's detached trunk latents (641 dimensions: the post-interaction vehicle, driver, environment, and side-vector states plus an automation-state bit) with a 41,153-parameter MLP, three forecasting seeds, and the same 979 training labels.
Its pooled gold AUROC on the 512-row layout is 0.589 with the ECAN trunk and 0.614 with the trunk of the Driver-WM-encoder variant, versus 0.712 for the current-state HGB probe on the same 512 anchors (0.716 on the full 509-anchor gold set).
The ECAN-latent-state probe is below the current-state probe on three of three seeds (mean $-0.124$); the other difference is inconclusive (one of three seeds), and the two latent-state probes do not differ (+0.026).
At validation-fitted thresholds the latent-state probes fire 1.3--1.9 times more often, and at load-matched thresholds they show no timing advantage over the current-state probe.
Trunk latents are trained for forecasting, not for the warning label; this diagnostic does not test a jointly trained warning objective, and the deployed head remains the current-state HGB.

An oracle diagnostic on the 512-row layout replaces predictions with true future features. The current-state base gives 0.712 pooled AUROC; true occupancy and recovery features give 0.722, and adding true onset scores gives 0.731. Predicted features give 0.708 in sample and 0.690 out of fold. The improvements over the current-state base have intervals containing zero. This restricted oracle is not an upper bound on every possible warning model. It tests one readout and one annotation target.

\subsection{Operating points and calibration}
At the load matched to the openpilot-based baseline on the 512-row layout, pooled warning recall is 0.460 versus 0.344, manual recall 0.581 versus 0.269, and engaged recall 0.344 versus 0.417. 
Precision and recall are measured on an enriched labeled set. The paired lead-time intervals include zero. At ten alerts/hour, the small number of detected gold positives prevents a stable low-load advantage claim.

On the same calibration layout, the warning probe's validation-temperature-scaled expected calibration error (ECE) is 0.028 and Brier score 0.103, compared with 0.408 and 0.273 for the openpilot-based baseline. Its temperature hits the search-grid boundary. Allowing a bias through Platt scaling narrows but does not remove the paired Brier advantage ($-0.016$, interval $[-0.027,-0.006]$). These results concern calibration to human warning labels in the sampled population, not calibration of every event-hazard chain or crash probability. A test-prevalence constant predictor has Brier score about 0.108, which provides scale for interpreting 0.103 but is not a deployable validation-fitted baseline.

\subsection{Event timing on automatically defined targets}
At an achieved load of 20 alerts/hour over the full test exposure, learned warning probes detect a greater share of warranted-condition events at least one second early than persistence (+0.055) and dwell extrapolation (+0.053). On sustained-demand events, a future-occupancy head exceeds the openpilot-based baseline by 0.067 but trails the dwell-timer rule by 0.202. These event families are automatically defined and should be kept separate from human judgments.

For road-demand onsets during driver inattention, the ECAN demand forecast detects 0.130 at least one second early versus 0.129 for a constant-velocity TTC rule at 20 alerts/hour. 

\subsection{Public feature pipelines}
Six frozen feature pipelines were evaluated with the same downstream warning-learning protocol: Intel OMZ gaze, Intel OMZ driver action, MediaPipe Face Landmarker, OpenFace 3.0, L2CS-Net, and VideoMAEv2.
Their pooled warning AUROCs are respectively 0.603, 0.533, 0.733, 0.657, 0.547, and 0.781, compared with 0.716 for our current-state probe.
Our head is above L2CS-Net and the two Intel pipelines, and inconclusive against MediaPipe, OpenFace, and VideoMAEv2.
The latter's higher point estimate is retained.

These are frozen feature extractors with newly trained warning probes, not deployed systems or faithful reproductions of each method's complete training recipe. Their larger backbone counts cannot be compared fairly with our small trainable head while excluding our own frozen perception stack. MediaPipe features added to our features improve engaged PR-AUC by 0.076 with interval $[0.014,0.187]$, without a supported improvement elsewhere. These comparisons do not establish general superiority over the evaluated feature pipelines.

\section{Offline human evaluation: design, deviations, and results}
\label{app:human}
\subsection{Frozen systems and operating points}
System A uses current-state and predicted-future features; B uses current-state, dwell, and face features and is the head deployed on the vehicle; C is the openpilot-based baseline.
A averages the three frozen model seeds, B is seed-invariant, and no system is retrained for this evaluation.
The sampling and analysis plan was frozen on September 12, 2026.
Annotation was completed on September 21.
The 299 clips are an offline evaluation of recorded driving, not a prospective participant trial.

Primary operating points use thresholds fitted on validation at 10 and 20 alerts/hour. Secondary points fit thresholds to achieve those loads on test, which is retrospective operating-point matching. A 60-second per-route cooldown applies throughout. The comparator has a score floor: its minimum test rate is 15.10/hour in manual driving and 15.66/hour in engaged driving. At ten/hour, its unthinned row is therefore a floor comparison, not a matched-budget result. Seeded random thinning over 200 replicates provides a sensitivity analysis, not an alternative deployed policy.

At the validation 20/hour point, achieved manual loads are 9.40/13.69/17.67 for A/B/C and engaged loads are 14.10/12.91/24.81. At the validation 10/hour point, A achieves 4.32/6.53 and B achieves 3.83/2.99 in manual/engaged driving. 

\subsection{Sampling and labels}
The test exposure is 28.50 manual hours and 25.10 engaged hours. An alert-union stratum (S1) oversamples disagreements between A, B, and C. A silent-background stratum (S2) samples times without nearby alerts. A separate model-independent set (S3) provides additional labeled clips but is excluded from natural-rate estimation. Regions within 12.5 seconds of prior gold anchors or earlier labeled clips are excluded from S1/S2. S1 merges alerts into events using a ten-second guard and allocates draws by system combination, automation state, and driver, subject to a per-driver cap. S2 samples moving rows farther than 12.5 seconds from these events.

S1 contains 159 rated clips after one clip without road video was excluded; S2 contains 40 and S3 contains 100. S1 coverage is 0.809 overall, with cell coverage from approximately 0.58 to 0.94 (Table~\ref{tab:coverage}). S2 coverage is 0.956 manual and 0.935 engaged. Drivers with no draw have zero inclusion probability. Post-stratification extends covered groups to their cells; this requires a representativeness assumption beyond ordinary inverse-probability weighting.

Each clip is shown in two stages. The prefix stage displays cabin, road, and bus history over $[t-5,t]$. Raters judge whether a warning is warranted and record visible behavior. Their responses are locked before the future $[t,t+10]$ is shown. The second stage records event type, timing, severity, and vehicle response. Warning-warrant labels and event labels are distinct: a clip can contain an event without warranting a warning at its anchor.

The completed store has 897 stage records for 299 clips, including 82 warranted, 215 not-warranted, and two unsure warning labels. The unsure labels are excluded from warrant-based analysis. Seventy-eight clips contain an event: 48 in S1, five in S2, and 25 in S3. Manual driving contributes 20 weighted-partition events from 11 drivers and 29 events across all strata; engaged driving contributes 33 and 49, respectively. The analysis uses one consensus label per clip, not 897 independent annotations.

\subsection{Estimators and uncertainty}
For S1/S2, weights are $w_i=(N_{\mathrm{cell}}/N_{\mathrm{covered}})/\pi_i$, where $\pi_i$ is the inclusion probability. Let $E_i$ indicate a labeled event, $o_i$ its onset, and $D_{Xi}$ indicate an alert from system $X$ in $[o_i-10,o_i+3]$. Event recall is the weighted ratio
\begin{equation}
\widehat R_X=\frac{\sum_i w_iE_iD_{Xi}}{\sum_i w_iE_i}.
\end{equation}
The original protocol calls this Horvitz--Thompson event recall; more precisely, it is a ratio of weighted total estimates. S3 is included only in separate unweighted event analyses and AUROC. Effective lead time averages onset minus first-alert time, assigning zero to undetected events. It therefore combines timing and detection coverage and is not simply lead time conditional on successful detection.

False alarms/hour use weighted alerts judged unwarranted divided by exposure. Precision is a weighted ratio of warranted alerts to all alerts. All paired comparisons use identical events and 2,000 shared driver-cluster bootstrap resamples. Cells with zero sampling coverage are not made design-unbiased by reweighting observed cells. The coverage table and weight concentration should be read with the aggregate estimates.

\subsection{Protocol deviations}
The planned two independent blinded raters and a separate outcome rater were replaced before analysis by a five-member discussion committee using one operator. Inter-rater agreement cannot be estimated. Outcome timing was not independently blinded to prefix judgments, although prefix labels remained locked. Consequently, the planned agreement thresholds are unavailable and timing may be affected by anchoring.

Further deviations include the unrenderable S1 clip and three route-end clips with shortened outcome windows. The minimum detectable effect (MDE) computed from the observed discordance is distinct from the registered calculation, which fixes discordance at 0.4.

\subsection{Results at all operating points}
Tables~\ref{tab:humanmanual} and \ref{tab:humanengaged} give all operating-point estimates. ``Val'' denotes validation-fitted thresholds and ``test'' denotes retrospective test-achieved loads. These targets are not the achieved validation-to-test rates reported above. C is shown at its floor when ten/hour is unattainable; the thinning sensitivity is separate.

\begin{table}[h]
\caption{Manual-driving human evaluation, 20 weighted-partition events. Point estimates are descriptive. Recall is weighted event recall; FA/h is estimated false alarms/hour. C's ten/hour rows use its attainable floor.}
\label{tab:humanmanual}
\centering\small
\begin{tabular}{llrrr}
\toprule
Operating point & System & Recall & FA/h & Precision\\
\midrule
Val 10 & A & .008 & 1.25 & .493\\
 & B & .064 & 1.05 & .478\\
 & C & .000 & 11.52 & .122\\
Val 20 & A & .233 & 4.03 & .513\\
 & B & .087 & 8.63 & .288\\
 & C & .000 & 12.41 & .175\\
Test 10 & A & .292 & 4.72 & .474\\
 & B & .064 & 3.21 & .441\\
 & C & .000 & 11.52 & .122\\
Test 20 & A & .326 & 10.02 & .441\\
 & B & .138 & 13.86 & .242\\
 & C & .000 & 14.08 & .209\\
\bottomrule
\end{tabular}
\end{table}

\begin{table}[h]
\caption{Engaged-driving human evaluation, 33 weighted-partition events. Large uncertainty reflects concentration of weight in a few background events. Parentheses show the reported 95\% interval for recall.}
\label{tab:humanengaged}
\centering\small
\begin{tabular}{llrrr}
\toprule
Operating point & System & Recall (95\% interval) & FA/h & Precision\\
\midrule
Val 10 & A & .409 (.00,.93) & 2.15 & .325\\
 & B & .000 (.00,.00) & .00 & 1.000\\
 & C & .409 (.00,.93) & 1.58 & .793\\
Val 20 & A & .437 (.00,.99) & 5.27 & .421\\
 & B & .410 (.00,.93) & 9.61 & .299\\
 & C & .000 (.00,.01) & 13.76 & .383\\
Test 10 & A & .437 (.00,.99) & 3.14 & .375\\
 & B & .000 (.00,.01) & 7.76 & .275\\
 & C & .409 (.00,.93) & 1.58 & .793\\
Test 20 & A & .002 (.00,.02) & 9.82 & .471\\
 & B & .410 (.00,.93) & 12.95 & .320\\
 & C & .000 (.00,.00) & 6.21 & .519\\
\bottomrule
\end{tabular}
\end{table}

At manual val20, recall intervals for A/B/C are $[.02,.54]$, $[.00,.42]$, and $[.00,.00]$. FA/h intervals are $[1.23,8.04]$, $[1.41,21.83]$, and $[2.96,29.31]$; precision intervals are $[.05,.85]$, $[.08,.78]$, and $[.02,.55]$. At manual test10, recall intervals are $[.06,.65]$, $[.00,.34]$, and $[.00,.00]$; at test20, $[.09,.72]$, $[.04,.61]$, and $[.00,.00]$. Zero-width empirical bootstrap intervals arise when the observed weighted events provide no variation in the statistic; they do not establish a population probability of exactly zero.

\begin{table}[h]
\caption{Paired manual-driving differences. Every row is descriptive under the registered fewer-than-60-events rule. An interval containing zero is inconclusive, not evidence of equivalence.}
\label{tab:humanpaired}
\centering\small
\begin{tabular}{llrr}
\toprule
Point & Pair & Recall difference [95\% interval] & Effective lead, seconds [interval]\\
\midrule
Val 20 & A$-$C & +.233 [.022,.543] & -.139 [-.452,-.019]\\
 & B$-$C & +.087 [.000,.422] & -.041 [-.205,.000]\\
 & A$-$B & +.146 [-.106,.413] & -.097 [-.331,.041]\\
Val 10 & A$-$B & -.056 [-.302,.000] & +.013 [-.013,.087]\\
\bottomrule
\end{tabular}
\end{table}

Table~\ref{tab:humanpaired} gives the paired manual-driving differences. The registered MDE for recall at 20 manual events is 0.40, exceeding the observed differences. The A$-$C recall interval excludes zero, but the protocol still classifies this small-event analysis as descriptive. Its effective lead-time difference is negative. A$-$B remains inconclusive on both operating points. These distinctions prevent a higher A$-$C recall estimate from being used as proof that future features improve the current-state probe or that alerts occur earlier.

Five silent-background events account for 0.997 of engaged event weight, with one carrying 0.425. One undetected manual event carries 0.313, and A's advantage over C depends on two A-only detected events with weights 0.146 and 0.023. Precision 1.000 for B at engaged val10 reflects a sparse weighted evaluation and does not establish error-free alerting. Thinning C to ten/hour leaves manual recall zero and gives engaged recall 0.245. The original floor comparison remains separately reported.

\begin{table}[h]
\caption{Secondary unweighted AUROC on the 299-clip set.
Rows differ across systems (139 for A and B versus 134 for C in manual driving), so these values are unpaired and are not the paper's headline warning result; the paired 509-anchor comparison is in Appendix~\ref{app:warning} and Section~\ref{sec:warning}.}
\centering\small
\begin{tabular}{llrr}
\toprule
Driving & System & Rows / positives & AUROC [95\% interval]\\
\midrule
Manual & A & 139 / 32 & .781 [.68,.87]\\
 & B & 139 / 32 & .750 [.65,.85]\\
 & C & 134 / 27 & .478 [.33,.70]\\
Engaged & A & 142 / 48 & .641 [.54,.74]\\
 & B & 142 / 48 & .654 [.55,.77]\\
 & C & 136 / 43 & .758 [.69,.83]\\
Pooled & A & 281 / 80 & .712 [.63,.79]\\
 & B & 281 / 80 & .708 [.63,.79]\\
 & C & 270 / 70 & .641 [.52,.75]\\
\bottomrule
\end{tabular}
\end{table}

\subsection{Coverage and sensitivity}
\begin{table}[h]
\caption{S1 coverage used in weighting. $N$ is the eligible cell count; covered denotes eligible events from drivers with positive sampling probability.}
\label{tab:coverage}
\centering\small
\begin{tabular}{llrrrr}
\toprule
Driving & Alert cell & $N$ & Covered & Coverage & Labeled\\
\midrule
Manual & A & 212 & 160 & .755 & 14\\
 & A+B & 169 & 119 & .704 & 12\\
 & A+B+C & 74 & 43 & .581 & 6\\
 & A+C & 56 & 33 & .589 & 10\\
 & B & 216 & 179 & .829 & 14\\
 & B+C & 62 & 44 & .710 & 10\\
 & C & 315 & 277 & .879 & 14\\
Engaged & A & 199 & 187 & .940 & 14\\
 & A+B & 150 & 128 & .853 & 11\\
 & A+B+C & 71 & 42 & .592 & 6\\
 & A+C & 46 & 32 & .696 & 10\\
 & B & 175 & 165 & .943 & 14\\
 & B+C & 82 & 66 & .805 & 10\\
 & C & 361 & 291 & .806 & 14\\
\bottomrule
\end{tabular}
\end{table}

Two independent re-derivations documented by the study team reproduced the labels, weights, detections, and 24 comparison decisions. Reproduction of the estimator does not remove sampling uncertainty, consensus-label limitations, or assumptions about uncovered groups.

The silent-time estimates are 191.2 warranted rows/hour in manual and 3,083.1 in engaged driving, with intervals $[0,518.4]$ and $[0,7549.8]$. Rows on a dense time grid are not independent behavioral episodes. These values therefore should not be described as that many missed warning events per hour. More labeled background episodes and a prospective study are needed before drawing conclusions about natural warning utility.

\section{Development history and statistical interpretation}
\label{app:provenance}
\subsection{Statistical interpretation}
BATON forecasting comparisons use 1,000 paired driver-cluster bootstrap resamples. Warning comparisons use 2,000; AIDE uses 10,000 paired clip resamples. Means and standard deviations across training seeds measure optimization variability. Bootstrap intervals measure sampling variability over drivers or clips for the specified predictions. These are distinct sources of uncertainty. Published baselines without per-example predictions cannot support a paired test.

Our criterion (Section~\ref{sec:setup}) combines same-sign intervals excluding zero on at least two of three seeds with an absolute mean difference exceeding its seed standard deviation. This is a consistency rule, not familywise error control. Many targets and variants are tested. Accordingly, we report the rule, effect sizes, failed comparisons, and model-development history rather than interpreting every qualifying cell as a separately confirmed discovery. A failure to detect a difference is inconclusive, not equivalence or demonstrated noninferiority. A lower NLL is better probabilistic fit; calibration requires additional analysis.


\subsection{Temporal scope and archive}
Temporal access is distinct from causal inference. An anchor cutoff applies to clip tokenization and upstream preprocessing as well as Transformer attention. Per-recording driver-position priors, history interpolation, pose reference updates, synchronization, and route-quantile target construction are potential sources of temporal leakage. Route-level target thresholds are retrospective benchmark definitions, rather than quantities available to an online anchor-state calculation. The method does not identify intervention effects, simulate counterfactual control policies, or recover a causal graph.

The experiment archive records configuration, normalization, checkpoint, code, and prediction-layout hashes. These identifiers distinguish data generations and align paired comparisons. Reproducing a result requires the corresponding configurations, predictions, and complete run manifests in addition to the summaries reported here.

\section{Road distillation and measured deployment}
\label{app:deployment}
\subsection{Student versions and training}
The student replaces the frozen V-JEPA~2 ViT-g clip encoder and the separate OpenBADAS risk chain. It produces the teacher's 1,408-dimensional pooled feature and a risk logit. The existing PCA-256 projection, token bank, visual side vector, joint model, HGB probe, and score thresholds remain fixed. The deployed checkpoint is v5; v6 and v7 are later offline variants.

Training teacher targets are cached at 0.5-second clip ends for pooled features and 1\,Hz for risk. Version v7 adds 16 spatial-region feature targets. The split uses 248 training routes (236 with cached features) and 46 validation routes; 158 test and 56 held-out routes are excluded from training. The full-data corpus contains 520,569 clips, yielding 511,780 training windows, with 2,728 validation windows. A clip contains 16 frames at $224\times224$ pixels with frame stride two. Six epochs use cosine learning-rate decay with a peak near $10^{-4}$. The v5 ViT-S run uses 10,323 steps per epoch; later full-data ViT-S and ViT-B runs use 16,380 and 32,651. Approximate epoch times are 3,900--4,000 seconds for ViT-S and 8,750 seconds for ViT-B on a workstation GPU.

The loss in Eq.~\eqref{eq:distill} combines feature reconstruction and risk matching. Region supervision is auxiliary: the deployed interfaces remain pooled features and risk. It improves fidelity without adding a region-token requirement to the live joint model. Loss coefficients and exact version manifests are part of the training configurations; the reported comparisons identify each version's changes rather than inferring a single training recipe from all versions.

\begin{table}[h]
\caption{Validation fidelity of the best checkpoint of each six-epoch student run. PCA $R^2$ is weighted by explained variance; side $R^2$ refers to the visual side-vector target. The deployed model is v5.}
\label{tab:studentfidelity}
\centering\small
\begin{tabular}{lrrrrr}
\toprule
Student & Raw cosine & Centered cosine & PCA $R^2$ & Mean PCA $R^2$ & Side $R^2$\\
\midrule
v5 ViT-S/16 & .979 & .886 & .796 & .281 & .400\\
v6 ViT-S/16, full data & .986 & .925 & .866 & .451 & .541\\
v7-S, region targets & .988 & .936 & .888 & .495 & .589\\
v7-B, region targets & .987 & .932 & .880 & .477 & .569\\
\midrule
Stated fidelity targets & .970 & -- & .900 & -- & .850\\
\bottomrule
\end{tabular}
\end{table}

All four versions pass the raw-cosine target; the projected-feature targets are not met (Table~\ref{tab:studentfidelity}), and ViT-B does not improve on ViT-S at equal epochs.

\begin{table}[h]
\caption{Token-bank fidelity over 60 routes and 299,398 anchors. Subscripts $w$ and $p$ denote variance-weighted and pooled estimates; the bar denotes a component mean. Availability agrees exactly because the flags depend on timing.}
\centering\small
\begin{tabular}{lrrrrr}
\toprule
Student & Token $R^2_w$ & Token $R^2_p$ & Side $\overline{R^2}$ & Side $R^2_p$ & Agreement\\
\midrule
v5 & .832 & .827 & .472 & .815 & 1.000\\
v6 & .856 & .851 & .534 & .840 & 1.000\\
v7-S & .872 & .868 & .578 & .857 & 1.000\\
v7-B & .863 & .859 & .552 & .848 & 1.000\\
\bottomrule
\end{tabular}
\end{table}
The full-bank fraction is 0.843 in both teacher and student replay layouts.
Live full-bank coverage is 0.0\% in drive 2 and 50.6\% in drive 3 (Table~\ref{tab:runtime}), against 0.843 in replay.
Timing agreement passes the 0.99 target.

\subsection{Replay comparison and isolated latency}
Four replay routes, two validation and two test, substitute both student features and student risk while retaining downstream components. Every evaluated student version reproduces the teacher-based alert-event set on these routes. Table~\ref{tab:replay} reports the deployed v5 student; later student versions reproduce the same event sets with 99th-percentile differences within 0.002 of these values. The current-state warning probe does not consume the forecast token output, so this diagnostic mainly tests the warning path, including the changed risk source.

\begin{table}[h]
\caption{Replay warning agreement on four routes (two validation, two test) with the deployed v5 student's road features and risk substituted for the teacher's. Score differences are the 99th percentile and maximum of $|\Delta|$; the median is zero on every route.}
\label{tab:replay}
\centering\small
\begin{tabular}{lrrrl}
\toprule
Route & Teacher / student events & q99 $|\Delta|$ & max $|\Delta|$ & Events\\
\midrule
R1 & 2 / 2 & .009 & .030 & Identical\\
R2 & 6 / 6 & .036 & .081 & Identical\\
R3 & 6 / 6 & .016 & .031 & Identical\\
R4 & 2 / 2 & .007 & .023 & Identical\\
\bottomrule
\end{tabular}
\end{table}

Isolated ViT-S latency with fp32 input is 101.3\,ms median and 109.1\,ms p95, including a 7.1\,MB upload. A trained uint8-input student uses 2.41\,MB and reaches 41.6--41.9\,ms median and 42.5--43.9\,ms p95. Approximately 16.2\,ms is compute and 25\,ms is upload and overhead. ViT-B takes 75.1\,ms median and 77.4\,ms p95. These isolated timings are distinct from the slower live student calls under concurrent load.

\subsection{Live configurations and adaptations}
The device uses a comma four processor with eight cores online and a USB-attached AMD RX 9060 GPU through tinygrad's AMD backend with fp16 execution. The openpilot driving model and driver monitor continue on the device processor. The GPU daemon schedules per-model priority slots and the orchestrator maintains the bus grid, histories, readiness, and warning output. Stale inputs trigger fallback to openpilot's driver monitor. No vehicle-control policy is learned or replaced by \model{}.

Three instrumented measurement drives last 1,347, 1,450, and 2,453 seconds; drive 1 (September 20) was driven by the experimenter and is not part of the study, and drives 2 and 3 (September 22) were driven by study participants with the experimenter in the passenger seat.
The first runs only the driver and vehicle paths.
The second adds a 2\,Hz road student at priority 2 and a nominal 0.2\,Hz heavy risk chain for logging.
The third disables that chain and raises the road student to priority 3.
Both latter drives use the student risk head with a ten-second validity window, 1\,Hz driver-box updates, and 2.5\,Hz compact/joint forecasting.
The warning probe ticks at 5\,Hz.
The 2.5\,Hz joint-model outputs are logged on every call; the 5\,Hz warning decision reads only the current-state features.

\begin{table}[h]
\caption{Achieved rates, module latencies, and input validity in three instrumented measurement drives. Drive 1 (September 20) was the experimenter's test drive and is not part of the study; drives 2 and 3 were driven by study participants on September 22. Drive 1 omits the road path, drive 2 retains a heavy risk chain for logging, and drive 3 uses the student alone. End-to-end latency runs from the driver-frame timestamp to the warning score; module latency is median/p95 in milliseconds; ``--'' marks road-path quantities that drive 1 did not run.}
\label{tab:runtime}
\centering\small
\begin{tabular}{lrrr}
\toprule
Quantity & Drive 1 & Drive 2 & Drive 3\\
\midrule
Duration (s) & 1,347 & 1,450 & 2,453\\
Pose / joint rate (Hz) & 9.67 / 4.61 & 8.50 / 2.26 & 9.38 / 2.35\\
Road-student rate (Hz) & -- & 0.93 & 1.89\\
Warning-readout rate (Hz) & 5.00 & 5.00 & 5.00\\
End-to-end latency p50 / p95 (ms) & 83 / 153 & 138 / 239 & 135 / 177\\
Ready ticks (\%) & 94.0 & 94.2 & 97.9\\
Full road-token bank (\%) & -- & 0.0 & 50.6\\
Valid visual side vector (\%) & -- & 19.3 & 93.1\\
\midrule
Pose latency & 48.9 / 61.5 & 53.5 / 71.3 & 50.5 / 66.9\\
Pose overwritten inputs (\%) & .4 & 9.0 & 3.9\\
Keypoint-forecaster latency & 23.9 / 33.7 & 29.6 / 40.3 & 26.4 / 38.1\\
Joint-model latency & 26.3 / 38.6 & 30.0 / 40.8 & 27.0 / 38.8\\
Road-student latency & -- & 76.5 / 94.6 & 70.5 / 90.9\\
Time to first ready (s) & 80.4 & 84.2 & 50.8\\
End-to-end p99 latency (ms) & 218 & 266 & 280\\
Pose validity (\%) & 96.3 & 88.2 & 94.6\\
Risk-margin validity (\%) & -- & 91.6 & 97.9\\
Median risk age (s) & -- & .9 & .4\\
\bottomrule
\end{tabular}
\end{table}

Table~\ref{tab:runtime} lists module latencies and input validity. In drive 2, the heavy risk chain completes only two clips in 1,450 seconds, while student cadence falls to 0.93\,Hz and the bank never fills. Drive 3 reaches 1.89\,Hz for the student, 9.38\,Hz for pose, and 2.35\,Hz for joint prediction. Its p95 latency of 177\,ms meets the 300\,ms target, and readiness of 97.9\% meets the 95\% target. Non-preemptive calls explain a scheduling constraint: a roughly 70\,ms live student or 96\,ms detector call does not fit entirely in the roughly 46\,ms gap between pose calls.

\subsection{Descriptive replay of the vehicle branch on study routes}
As a descriptive check that the logged forecasting backbone produces usable predictions on the study drives, we replayed the vehicle branch alone on 5.0 hours of CAN from the 13 study routes on which \model{} presented warnings.
For the probability of a first onset within 3\,s, AUROC is 0.907 for steering-phase onset, 0.798 for response onset, 0.543 for brake press, and 0.593 for gas release, against labels derived from the same CAN signals with fixed thresholds.
Predicted steering-onset probability rises monotonically toward events, from 0.19 at 5--10\,s before onset to 0.64 within 1\,s.
At the anchors where a warning was presented, the predicted steering-onset probability is elevated (median 0.174 versus 0.088 over all anchors).
These are single-seed, vehicle-only replays with self-derived labels and autocorrelated anchors, without driver-clustered intervals; they describe the logged backbone and do not bear on the warning decision, which reads current states only.

\section{Completed on-road study: questionnaire and results}
\label{app:userstudy}
\subsection{Procedure}
Questionnaire sessions were held on September 22 and 23, 2026, with analysis on September 23; recording began on September 21.
Fourteen participants completed a within-participant comparison of deployed \model{} and openpilot's driver-monitoring system in their own compatible vehicles: twelve on the first day and two on the second.
Recordings of the two devices over September 21--23 total 8.4 hours (504 minutes over 39 loggerd routes), 5.9 hours on the \model{} device and 2.5 hours on the openpilot device, computed from the segment videos.
Participants drove in every recorded session on September 21--23, including the instrumented measurement drives of September 22; the experimenter was always in the passenger seat.

Condition order alternated between participants and is documented in timestamped video recordings of every drive and questionnaire session; system identity was concealed from participants.
Twelve PDFs contained readable form fields and two were read from hand-circled responses.

\subsection{Questionnaire and scoring}
For each condition, responses use 1 = strongly disagree through 7 = strongly agree, with 4 neutral and a separate ``Cannot judge'' option. The five statements are:
\begin{enumerate}
\item The driver-monitoring warnings during this drive fit the driving situations at the time.
\item The driver-monitoring warnings during this drive were necessary for me.
\item Overall, the driver-monitoring warnings during this drive were appropriate.
\item The driver-monitoring warnings during this drive came at the right time.
\item The driver-monitoring warnings during this drive annoyed me.
\end{enumerate}
After both drives, participants select a preferred condition or ``No difference'' and may provide a free-text comment. These repeated five-item blocks plus two final questions constitute the 12-question form. Appropriateness is the mean of available items 1--3; timeliness is item 4; annoyance is item 5. Item 3 is not reverse-scored. ``Cannot judge'' and unanswered responses are missing. The primary paired analysis uses available construct pairs rather than treating the two drives as independent participants.

\begin{table}[h]
\caption{Participant-level construct scores. C is \model{}; S is openpilot's driver-monitoring system (stock system). Missing values are shown as dashes. Preference is C, S, or equal.}
\label{tab:participants}
\centering\small
\begin{tabular}{lrrrrrrl}
\toprule
ID & APP C & APP S & TIM C & TIM S & ANN C & ANN S & Preference\\
\midrule
P01 & 4.00 & 2.50 & 7 & 1 & 6 & -- & S\\
P02 & 6.33 & 4.00 & 7 & 1 & 2 & 2 & C\\
P03 & 4.00 & 1.67 & 4 & 2 & 2 & 1 & C\\
P04 & 6.33 & 1.67 & 6 & -- & 3 & -- & C\\
P05 & 6.00 & 5.00 & 7 & 2 & 4 & 4 & C\\
P06 & 4.33 & 3.67 & 6 & 4 & 5 & 1 & S\\
P07 & 4.00 & 4.00 & 4 & 4 & 1 & 1 & Equal\\
P08 & 3.33 & 3.00 & 4 & 3 & 5 & 1 & S\\
P09 & 6.67 & 1.67 & 6 & 1 & 1 & 1 & C\\
P10 & 5.67 & 1.67 & 6 & 1 & 2 & 1 & C\\
P11 & 7.00 & 7.00 & 7 & 7 & 1 & 1 & Equal\\
P12 & 3.67 & 3.67 & 5 & 5 & 2 & 1 & S\\
P13 & 6.00 & 6.67 & 6 & 6 & 1 & 1 & Equal\\
P14 & 6.50 & 2.67 & -- & 3 & 5 & 2 & C\\
\bottomrule
\end{tabular}
\end{table}

\subsection{Uncertainty, missingness, and preference}
The reported uncertainty uses 2,000 participant-level paired bootstrap resamples with seed 20260922 and percentile 95\% intervals, alongside two-sided Wilcoxon signed-rank tests. The primary appropriateness endpoint has 14 pairs. Timeliness has 12 because P04 could not judge the timeliness of openpilot's warnings and P14 left the \model{} timing item unanswered. Annoyance has 12 because P01 could not judge annoyance with openpilot's warnings and P04 omitted that response. The annoyance means in Figure~\ref{fig:userstudy} refer to those twelve paired participants, not to all available responses under each condition separately.

The primary mean differences are 1.79, 2.67, and 1.17 for appropriateness, timeliness, and annoyance. Median differences are 1.25, 2.00, and 0.50. The Wilcoxon values are .008, .008, and .031. Secondary endpoints are not treated as independent confirmatory discoveries. For the twelve first-day participants, the mean differences are 1.82 $[.87,2.88]$, 2.91 $[1.55,4.27]$, and 1.10 $[.30,2.20]$, with preference 6 C, 4 S, and 2 equal.

Neutral imputation for missing endpoint scores yields mean differences 1.79/2.50/1.07, with intervals $[.86,2.75]$, $[1.36,3.71]$, and $[.36,1.93]$. The corresponding reported $p$ values are .008/.005/.028. This sensitivity retains the available-item appropriateness composites; it is not imputation of every missing questionnaire item before averaging.

Preference counts are 7 C, 4 S, and 3 equal. The respective shares and Wilson intervals are .50 $[.27,.73]$, .29 $[.12,.55]$, and .21 $[.08,.48]$. The bootstrapped preference-share difference C minus S is .21 $[-.21,.64]$. It supports a numerical tilt rather than a resolved population preference.

\subsection{Comments and interpretation}
Comments describe \model{} as quick, accurate, and able to respond to traffic situations such as pedestrians crossing and vehicles stopping. Others mention frequent warnings, loud sound, uncertainty about the reason for a warning, or a preference for a symbol and danger-coded colors over text that a non-native speaker cannot read in time. Comments on openpilot's driver-monitoring system include missed or late warnings and a perceived focus on hands-off behavior. These are participant descriptions of their experience, not specifications of the sensing capabilities of openpilot's driver-monitoring system. We do not infer from them that the deployed research system detects every described hazard correctly.

Ten participants give higher appropriateness to C, three give equal composite scores, and one (P13) gives S a higher score with both composites above six and an equal preference.
Eight of twelve complete timing pairs favor C and four are equal.
Annoyance is higher for six of twelve complete pairs and equal for six.
Nine of twelve comparable openpilot annoyance scores equal one, while eight of thirteen openpilot timeliness scores are at most three.

The comments were summarized descriptively. They suggest requiring valid pose inputs and sustained score excursions before a warning and varying acoustic intensity, all with fixed learned models and thresholds; this study does not test these changes.

\end{document}